\documentclass[11pt]{article}
\usepackage{times}
\usepackage{latexsym}

\pdfoutput=1

\PassOptionsToPackage{prologue,svgnames}{xcolor}
\usepackage[final]{acl}
\usepackage{xcolor, colortbl}
\usepackage[svgnames]{xcolor}

\usepackage{booktabs}
\usepackage{multirow}

\usepackage{graphicx} 
\usepackage{float}
\usepackage{subfigure} 
\usepackage{amssymb}

\usepackage{hyperref}
\usepackage{tablefootnote}
\usepackage{xspace}

\usepackage{float}
\usepackage{amsmath}
\usepackage{bm}
\usepackage{algorithmicx}
\usepackage{algorithm}
\usepackage{algpseudocode}
\usepackage{fontawesome}

\usepackage{arydshln}

\usepackage{amssymb}
\usepackage{pifont}
\newcommand{\cmark}{\ding{51}}%
\newcommand{\xmark}{\ding{55}\xspace}%
\usepackage{enumitem}

\usepackage{microtype}
\usepackage{hyperref}
\usepackage{url}
\usepackage{amssymb}
\usepackage{pifont}
\usepackage{booktabs}
\usepackage[utf8]{inputenc}
\usepackage{graphicx}
\usepackage{subcaption}
\usepackage{microtype}
\usepackage{inconsolata}
\usepackage{boxedminipage}
\usepackage{amsmath}
\usepackage{xspace}
\usepackage{arydshln}
\usepackage{multirow}
\usepackage{multicol}
\usepackage{xcolor}
\usepackage{soul}
\definecolor{lavender}{rgb}{0.94, 0.92, 0.95}
\usepackage[T1]{fontenc}
\addtocontents{toc}{\protect\setcounter{tocdepth}{-1}}

\newcommand{\data}{\textsc{M3SciQA}\xspace} 
\newcommand{\datanum}{\textsc{1,452}\xspace}
\newcommand{\bb}[1]{\textbf{#1}}

\usepackage{skak}

\newcommand{\aspace}{\hspace{1em}}
\newcommand{\equalCon}{$^*$}

\newcommand{\yale}{%
    $\raisebox{1mm}
    {\includegraphics[width=0.018\textwidth]{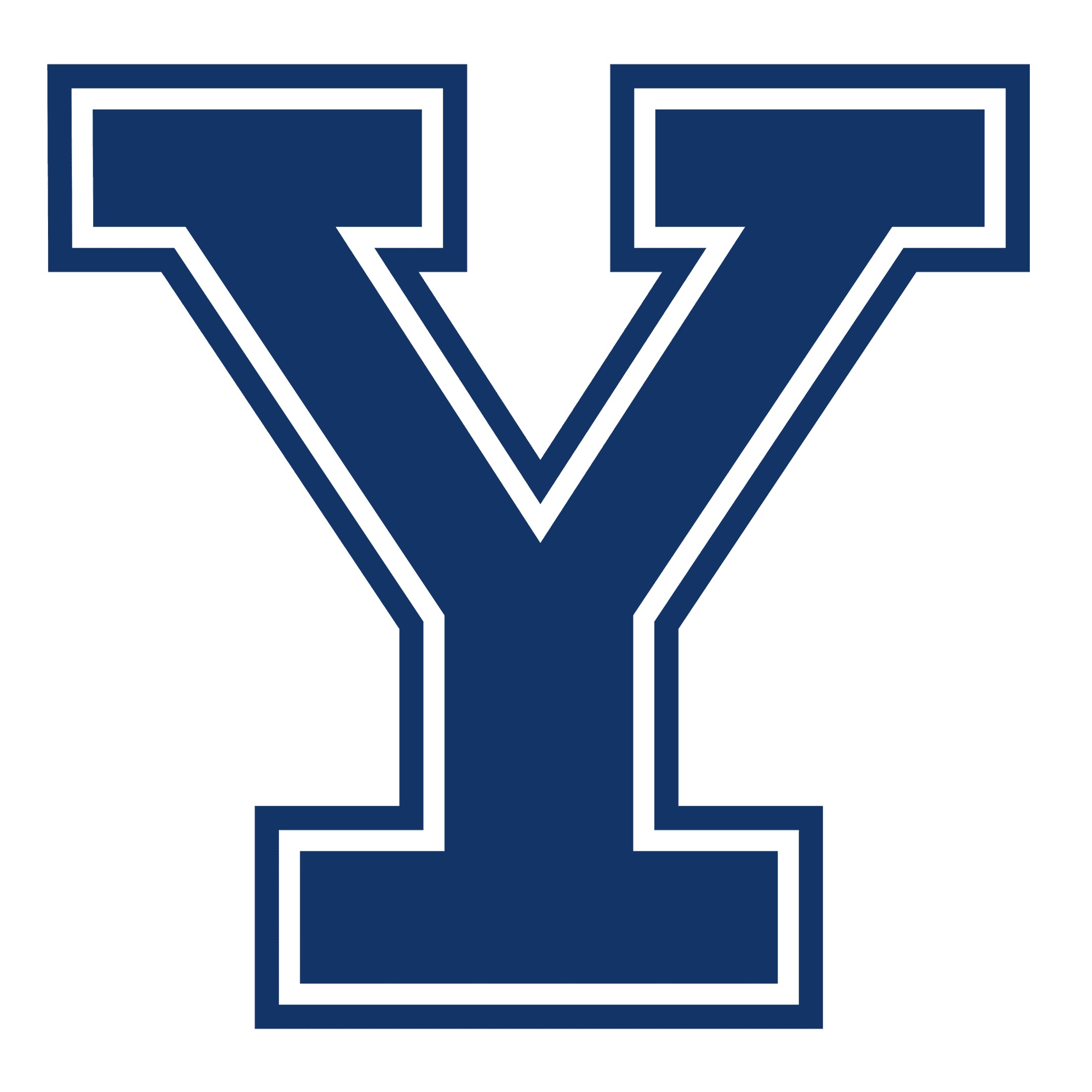}}$ 
}
\newcommand{\aiTwo}{%
    $\raisebox{1mm}
    {\includegraphics[width=0.018\textwidth]{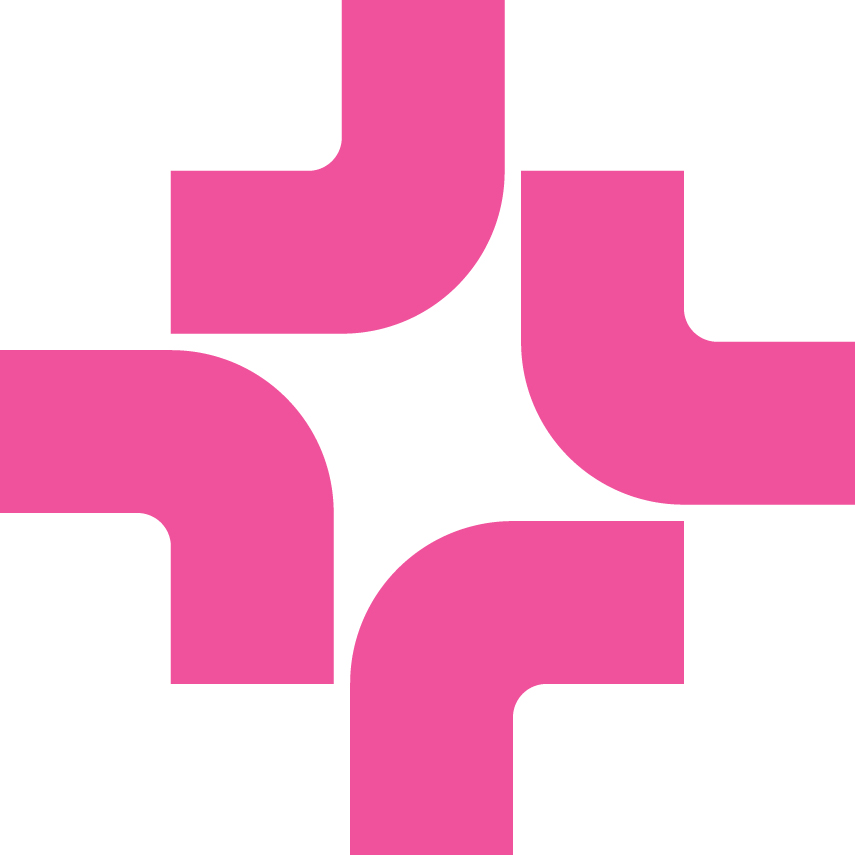}}$ 
}

\title{\data: A Multi-Modal Multi-Document Scientific QA\\ Benchmark for Evaluating Foundation Models}

\author{
    Chuhan Li \yale\thanks{Equal contribution.} \aspace 
    Ziyao Shangguan \yale\equalCon \aspace 
    Yilun Zhao \yale  \aspace 
    Deyuan Li \yale  \aspace \\
    \textbf{Yixin Liu} \yale  \aspace 
    \textbf{Arman Cohan} \yale \aiTwo \aspace\\
    \yale Yale University \aspace
    \aiTwo Allen Institute for AI \\
      \texttt{\{chuhan.li.cl2575, ziyao.shangguan\}@yale.edu}
}

\newcommand{\eg}{\hbox{\emph{e.g.,}}\xspace}

\newcommand{\wrt}{\hbox{\emph{w.r.t.}}\xspace}
\definecolor{Gray}{gray}{0.95}
\newcolumntype{a}{>{\columncolor{Gray}}c}
\begin{document}
\maketitle

\begin{minipage}[t]{2\linewidth}
\vspace{-1.3cm}
  \centering
  \href{https://github.com/yale-nlp/M3SciQA}{{\faGithub{}}\xspace\texttt{https://github.com/yale-nlp/M3SciQA}} 
\end{minipage}

\begin{abstract}
    
Existing benchmarks for evaluating foundation models mainly focus on single-document, text-only tasks.
However, they often fail to fully capture the complexity of research workflows, which typically involve interpreting non-textual data and gathering information across multiple documents.
To address this gap, we introduce \data, a multi-modal, multi-document scientific question answering benchmark designed for a more comprehensive evaluation of foundation models. 
\data consists of \datanum expert-annotated questions spanning 70 natural language processing paper clusters, where each cluster represents a primary paper along with all its cited documents, mirroring the workflow of comprehending a single paper by requiring \emph{multi-modal} and \emph{multi-document} data. 
With \data, we conduct a comprehensive evaluation of 18 foundation models. 
Our results indicate that current foundation models still significantly underperform compared to human experts in multi-modal information retrieval and in reasoning across multiple scientific documents. 
Additionally, we explore the implications of these findings for the future advancement of applying foundation models in multi-modal scientific literature analysis.

\end{abstract}
\definecolor{reference}{RGB}{204, 240, 187}  
\definecolor{anchor}{RGB}{255, 229, 152} 

\definecolor{locality}{RGB}{248, 206, 204} 
\definecolor{detail}{RGB}{211, 225, 245}

\newcommand{\hlreference}[1]{{\sethlcolor{reference}\hl{#1}}}
\newcommand{\hlanchor}[1]{{\sethlcolor{anchor}\hl{#1}}}
\newcommand{\hllocality}[1]{{\sethlcolor{locality}\hl{#1}}}
\newcommand{\hldetail}[1]{{\sethlcolor{detail}\hl{#1}}}
\begin{figure*}[!hbt]
\centering
\includegraphics[width=1\textwidth]{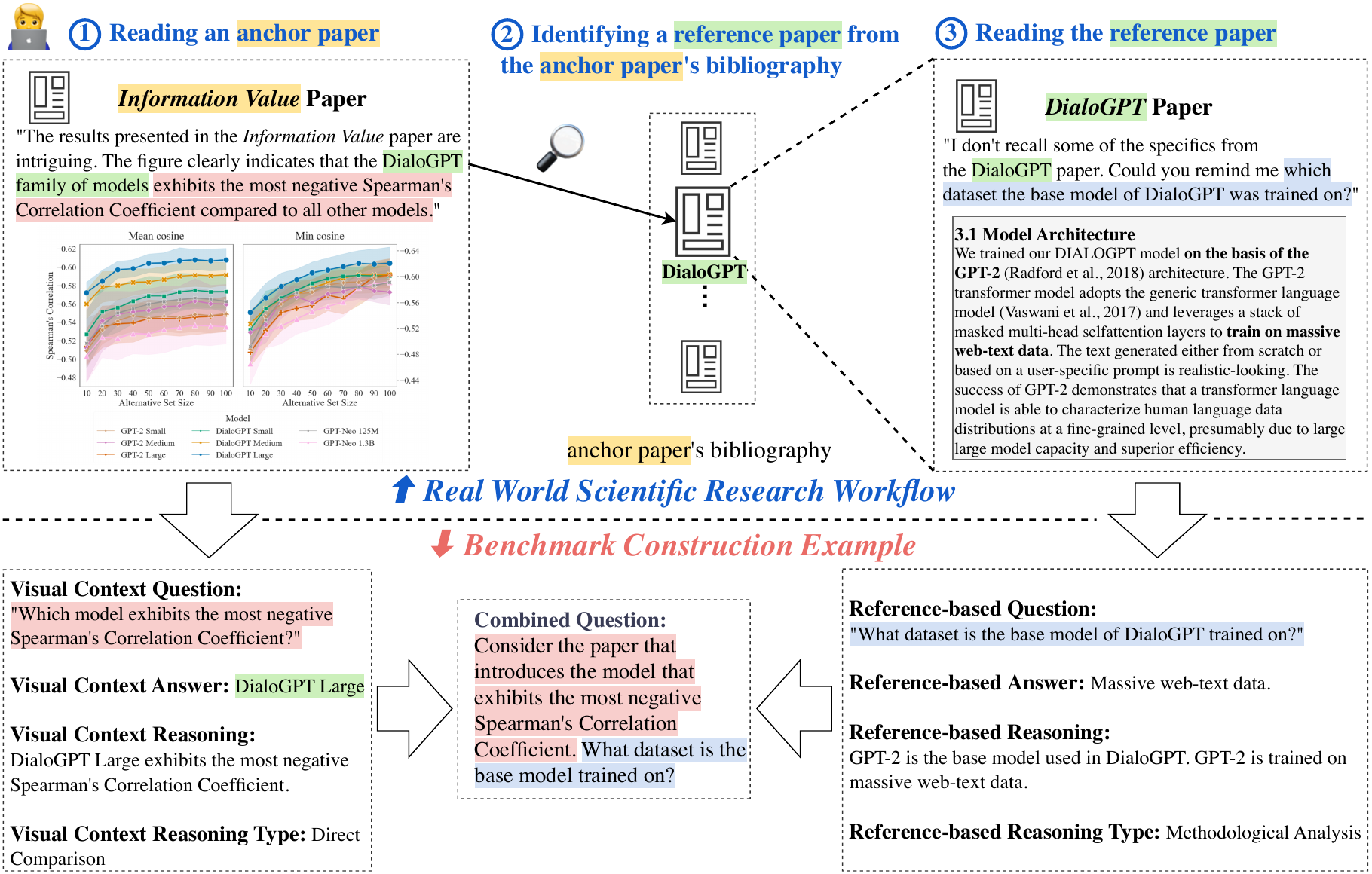}
\caption{
(\textbf{Top}) The common workflow of comparative analysis in scientific research, particularly when a result, such as a figure/table in the \hlanchor{Information Value paper (anchor paper)} \citep{giulianelli2023information}, prompts further examination of related research, such as details from \hlreference{DialoGPT (reference paper)} \citep{zhang2020dialogpt}. \textbf{(Bottom)} A demonstration of the workflow for constructing a \hllocality{visual context question}, \hldetail{reference-based question}, and combined question.
}

\label{fig:real_world}
\end{figure*}

\section{Introduction}

In scientific research, the findings presented in a paper often serve as a foundation for further investigation. When studying research papers, researchers typically explore related and cited scholarly works to acquire additional context and insights. Simultaneously, research papers are inherently multi-modal, presenting additional and often important insights in the form of figures and tables. Such properties can pose challenges for AI systems in accurately interpreting and integrating diverse data formats across multiple research papers.

Recent studies have showcased foundation models' remarkable performance across a variety of tasks in scientific literature understanding, including summarization~\citep{goyal2023news, liu2023benchmarking}, document-based question answering~\citep{newman-etal-2023-question, zhao-etal-2024-docmath, xu2024kiwi}, and scientific figure question answering~\citep{masry2022chartqa, yue2023mmmu, lu2024mathvista}.
However, current investigations are mostly confined to a \emph{single-document} or \emph{text-only} setting, ignoring the \emph{multi-modal} and \emph{multi-document} nature of scientific research, where insights are often derived from interpreting interconnected texts, figures, and tables across multiple scholarly works. 

To address this gap, we introduce \data, a \underline{\textbf{M}}ulti-\underline{\textbf{M}}odal, \underline{\textbf{M}}ulti-document \underline{\textbf{Sci}}entific \underline{\textbf{Q}}uestion \underline{\textbf{A}}nswering benchmark. 
This benchmark contains \datanum expert-annotated questions spanning 70 natural language processing (NLP) paper clusters, encompassing 3,066 papers. 
Each paper cluster comprises of an anchor paper and all its cited papers.
Inspired by the common workflow of comparative analysis in scientific research (as illustrated in Figure~\ref{fig:real_world}), our benchmark simulates a process in which a finding, derived from a \emph{scientific image} in the anchor paper, 
prompts further investigation into a specific referenced paper.
This simulation enriches the benchmark by requiring the models to engage in \emph{cross-referencing} among related documents, setting a new testbed for evaluating foundation models in scientific documents understanding and reasoning (Section \ref{sec:overview}). 

We evaluate a wide spectrum of \emph{open-source} and \emph{proprietary} large language models (LLMs) and large multi-modal models (LMMs). 
Our experimental results reveal significant limitations in both open-source and proprietary LMMs, particularly in their ability to translate and interpret scientific images and perform effective re-ranking based on these images, with the best-performing model, GPT-4o, achieving a Mean Reciprocal Rank (MRR) of 0.488 compared to a human expert score of 0.796, corresponding to a performance gap of 0.308. 

Similarly, we observe that both open-source and proprietary LLMs struggle with long-range retrieval tasks,
specifically with extracting and analyzing information from one or more academic documents. Here, the best-performing model, Command R+, achieves an accuracy score of 33.25 compared to an human expert accuracy score of 76.56\footnote{Human expert performance is assessed in the setting where the correct reference paper is known.}.
These findings underscore the challenges that current models face in handling complex, \emph{multi-modal}, \emph{multi-document}, and domain-specific information.

Our main contributions are as follows:
\vspace{-0.5em}
\begin{itemize} [leftmargin=*]
\itemsep0em 
\item We introduce \data, a comprehensive benchmark designed to evaluate the multi-modal reasoning abilities in interpreting multiple scientific documents. 
\item We conduct an extensive evaluation covering a wide range of LMMs and LLMs. Our experimental results reveal a noticeable performance gap between foundation models and human experts. 
\item To better understand the limitations of current foundation models, we conduct a detailed analysis of scientific figure information retrieval, long-context re-ranking, and long-range retrieval, providing valuable insights for future advancements of foundation models.
\end{itemize}

\begin{figure*}[hbt!]
\centering
\includegraphics[width=1\textwidth]{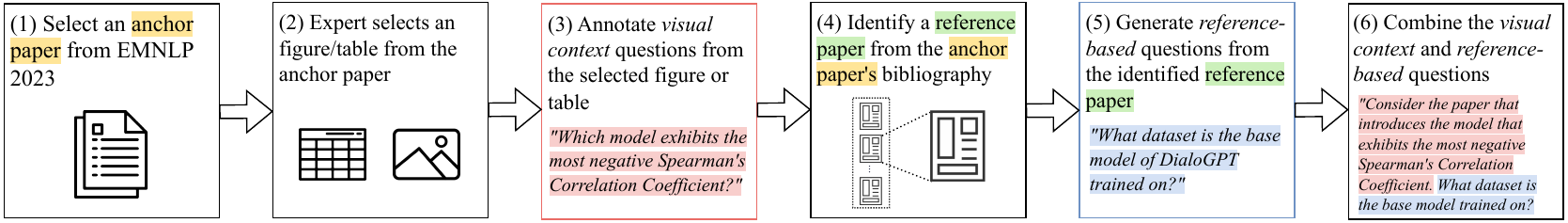}
\caption{
An overview of \data question construction pipeline.
}
\label{fig:pipeline}
\end{figure*}
\section{The \data Benchmark} 
\subsection{Overview of \data}
\label{sec:overview}

\begin{figure*}[!ht]
    \centering
    \includegraphics[width=\textwidth]{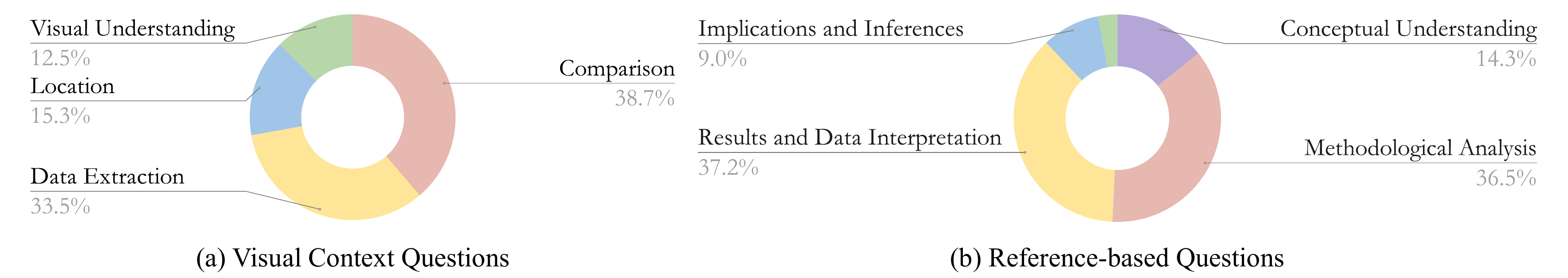}
    \caption{Distribution of reasoning types of \emph{visual context} and \emph{reference-based} questions in \data.} 
    \label{fig:reasoning_distribution}
\end{figure*}

Our objective is to develop a challenging yet realistic QA benchmark that necessitates both \emph{multi-modal} and \emph{multi-document} reasoning over scientific papers. 
%
%
To achieve this objective, we define two types of intermediate questions in our question construction pipline: 
\vspace{-0.2em}
\begin{itemize}[leftmargin=*]
    \itemsep0em 
    \item \textbf{Visual Context Question}: A question derived from a figure or table of an \textit{anchor paper}, with its answer pointing to a \textit{reference paper}. Each figure or table can correspond to multiple visual context questions.
    \item \textbf{Reference-based Question}: A question regarding a specific detail in the \textit{reference paper}. Each visual context question can correspond to multiple reference-based questions. 
\end{itemize}
The final \textit{combined questions} are created by combining each visual context question with each of its related reference-based questions.
The overview of this pipeline is shown in Figure \ref{fig:pipeline}. 
In constructing \data, expert annotators are tasked with composing visual context questions from the 70 curated anchor papers, adhering to four pre-defined reasoning categories: \emph{comparisons}, \emph{data extraction}, \emph{locations}, and \emph{visual understanding} (Table \ref{tab:locality_reasoning} in Appendix~\ref{app:locality_reasoning_examples}).
By answering a visual context question, expert annotators can pinpoint a reference paper that provides further elaboration on the topic from among all the publications cited by the anchor paper. 
Subsequently, GPT-4\footnote{\texttt{gpt-4-0125-preview}}
is employed to generate reference-based questions from the identified reference paper. 
GPT-4 is utilized again to rephrase and combine each visual context question with each of the related reference-based questions to form a comprehensive question that embodies both multi-modal and multi-document reasoning.
Finally, expert annotators are tasked with verifying the quality of these GPT-4-assisted questions.
Statistics of the benchmark are listed in Table \ref{tab:dataset_statistics}; 
distributions of reasoning types across visual context and reference-based questions are illustrated in Figure \ref{fig:reasoning_distribution}.

\subsection{Benchmark Construction Principles}

To bridge the gap in current benchmarks that separately assess either multi-modal or multi-document reasoning, our benchmark, \data, aims to encompass both elements in a single QA pair. 
Therefore, our benchmark construction pipeline adheres to the following guidelines:
(1) it includes diverse modalities, such as texts, figures (including line plots, bar plots, scatter plots, etc.), and tables (stored as images to preserve format integrity rather than as plain texts); 
(2) it necessitates connecting information across multiple documents; 
(3) it spans a variety of reasoning types, including four types of visual context reasoning and five types of reference-based reasoning;
(4) it poses significant challenges in both \emph{multi-modal} comprehension and \emph{multi-document} information retrieval; and 
(5) it generates realistic QA pairs that reflect the workflows common in scientific literature analysis.

\subsection{Benchmark Construction}
\label{sec:benchmark_construction}

\paragraph{Expert Annotators.}
We recruit three computer science graduate students with expertise in the field of NLP, each of whom have authored at least one peer-reviewed publication in top-tier NLP conferences. 
Their responsibilities include: 
(1) curating anchor papers from a pool of candidates and composing visual context questions; 
(2) reviewing and verifying the reasoning types of reference-based questions; 
(3) resolving discrepancies between answers generated from the two rounds of reference-based answer generation; 
and (4) checking consistency, clarity, and redundancy in the combined questions. 
Further details on annotations are provided in Appendix \ref{sec:expert_annotation_details}.

\paragraph{Anchor Papers.} 
To mitigate the risk of data contamination, where models might rely on pre-trained knowledge to answer the visual context questions rather than analyzing the provided scientific images, we curate anchor papers from a recent NLP conference, EMNLP 2023. 
Among the 1,047 papers accepted by EMNLP 2023, we select 441 papers that were released on arXiv after October 1st, 2023 as candidate anchor papers. 

\paragraph{Visual Context QAs from Anchor Papers.} 
Two of the expert annotators curate 70 papers by manually examining 441 candidate anchor papers collected. Subsequently, they select 21 figures and 62 tables from the 70 papers to compose 300 visual context questions and answers that conform to four visual reasoning types.
The ground truth answer to each visual context question is the single reference paper to which the visual context question directly refers.
This facilitates a transition from an anchor paper to a reference paper that elaborates on the subject.
The third annotator is responsible for validating the accuracy and relevance of these questions and answers.
371 papers are excluded in this process because they either lack figures or tables that can be analyzed by one of the reasoning types, or transition to a cited paper that is not available on arXiv.
Furthermore, due to the occurrence of identical answers among some visual context questions, these 300 questions correspond to only 107 reference papers.

\paragraph{Reference-Based QAs from Reference Papers.} 
By requiring that the 107 reference papers be available on arXiv, we ensure access to their complete content.
This enables us to utilize GPT-4 to generate open-ended, reference-based questions from the papers.
For each reference paper, we create five questions each corresponding to a reasoning type illustrated in Table \ref{tab:detail_reasoning_definition} in \ref{app:detail_reasoning_def}. 
These questions are designed to be answerable in a text-only setting, without the need for visual reasoning or OCR. 
Considering the possibility that GPT-4 may incorrectly categorize the questions, expert annotators manually examine the reasoning types associated with the questions and reassign when necessary. 
This process yields a total of 519 reference-based questions after filtering out duplicates, overly complex questions, questions that do not require specific insights from the paper (\eg ``What is the mathematical expression for calculating the F-1 score?''), and questions that do not belong to any of the five predefined reasoning types.
To establish a \emph{gold answer} for each question, we generate answers in two rounds. 
In the first round, answers are generated concurrently with the questions. 
In the second round, the model is prompted to answer the questions using the reference paper as context. 
We employ GPT-4 to determine whether the answers from both rounds are consistent. 
If any discrepancy is identified, expert annotators are enlisted to verify and finalize the answers. 
Further details can be found in Appendix \ref{2_round_checking}.

\paragraph{Combined Questions.} 
We utilize GPT-4 to compile the final questions for the benchmark by combining each visual context question with its corresponding reference-based questions. 
After the combination, expert annotators are tasked with verifying the question validity and rephrasing the GPT-4-assisted combined question when necessary.
Overall, we form \datanum combined questions, each associated with a specific figure or table. 
Expert annotators then review these combined questions to ensure that each visual context question and its corresponding reference-based questions are logically connected and relevant. 
They also check for clarity, consistency, and redundancy to maintain the quality and difficulty of the benchmark.

\begin{table}[!t]
\centering
\small
    \begin{tabular}{lr}
        \toprule
        \bb{Statistics} & \bb{\emph{Avg.} Value}   \\ 
        \midrule
        Visual Context Question Length \emph{(tokens)} & 12.9 \\
        Reference-based Question Length \emph{(tokens)} & 25.95 \\
        Combined Question Length \emph{(tokens)}  & 41.3 \\
        Answer Length \emph{(tokens)} & 24.9\\
        \addlinespace[0.2em]\hdashline\addlinespace[0.2em]
        \# Cluster & 70 \\
        \# Anchor Paper per Cluster & 1 \\
        \# Reference Paper per Cluster & 42.8\\
        Paper Length \emph{(tokens)} & 11.8K \\
        \midrule
        Validation Set Size  & 452 \\
        Test Set Size & 1000 \\
    
        \bottomrule
    \end{tabular}
    \caption{Key statistics of the \data benchmark.}
    \label{tab:dataset_statistics}
\end{table}

\begin{table*}
  \centering
  \small
  \begin{tabular}{llcccccccac@{}}
    \toprule
    \multirow{4}{*}{\vspace{0em}\textbf{Model}} 
    & \phantom{} 
    & \multicolumn{2}{c}{\textbf{Modality}} 
    & \phantom{}
    & \multicolumn{4}{c}{\textbf{Reasoning Type}} 
    & \cellcolor{white}
    & \phantom{}\\
    
    \cmidrule{3-4} \cmidrule{6-9} 

    \phantom{} 
    & \phantom{} 
    & \textbf{Table} 
    & \textbf{Figure}   
    & \phantom{}  
    & \textbf{COM} 
    & \textbf{DE} 
    & \textbf{LOC} 
    & \textbf{VU} 
    & \textbf{All} \\

\midrule
Expert Performance & & 0.678 & 0.765 & & 0.751 & 0.872 & 0.711 & 0.732 & 0.796\\
Random & & 0.134 & 0.106 & & 0.134 & 0.130 & 0.110 & 0.111 & 0.126 \\
\midrule
\multicolumn{11}{c}{\textbf{Simple Baselines}}  \\ 
\midrule
\texttt{text-embedding-3-large} & & 0.321 & 0.239 & & 0.267 & 0.323 & 0.384 & 0.218 & 0.297\\
\texttt{text-embedding-3-small} & & 0.223 & 0.205 & & 0.221 & 0.223 & 0.267 & 0.138 & 0.217\\
\texttt{text-embedding-ada-002} & & 0.185 & 0.168 & & 0.200 & 0.171 & 0.224 & 0.096 & 0.180\\
Contriever & & 0.165 & 0.229 & & 0.196 & 0.144 & 0.274 & 0.142 & 0.184\\
BM25 & & 0.138 & 0.098 & & 0.118 & 0.128 & 0.160 & 0.110 & 0.127\\
\midrule
\multicolumn{11}{c}{\textbf{Open-Source Large Multi-modal Models (LMMs)}} \\ 
\midrule

InternVL-Chat-V1.1 & & 0.168 & 0.084 & & 0.136 & 0.153 & 0.170 & 0.109 & 0.144\\
Yi-VL-34B & & 0.105 & 0.057 & & 0.101 & 0.088 & 0.080 & 0.086 & 0.091\\
Qwen-VL-Plus & & 0.065 & 0.131 & & 0.077& 0.053 & 0.148 & 0.136 & 0.089\\
LLaVA-1.6 & & 0.079 & 0.000 & & 0.088 & 0.044 & 0.052 & 0.000 & 0.056\\
DeepSeek-VL & & 0.075 & 0.087 & & 0.064 & 0.081 & 0.109 & 0.070 & 0.079\\
%
\midrule
\multicolumn{11}{c}{\textbf{Proprietary Large Multi-modal Models (LMMs)}} \\ 
\midrule
GPT-4o & & \bb{0.520} & \bb{0.454} & & \bb{0.443} & \bb{0.565} & \bb{0.570} & \underline{0.418} & \bb{0.500}\\
GPT-4V(ision) & & \underline{0.440} & 0.309 & & \underline{0.383} & \underline{0.407} & \underline{0.523} & 0.288  & \underline{0.400}\\
Claude-3-Sonnet & & 0.385 & \underline{0.369} & & 0.357 & 0.363 & 0.395 & \bb{0.422} & 0.374\\
Claude-3-Opus & & 0.256 & 0.343 & & 0.320 & 0.362 & 0.301 & 0.204 & 0.316\\
Gemini-Pro-Vision-1.0 & & 0.217 & 0.188 & & 0.196 & 0.160 & 0.284 & 0.195 & 0.197\\
Claude-3-Haiku & & 0.189 & 0.188 & & 0.194 & 0.201 & 0.130 & 0.208 & 0.188\\
\bottomrule
\end{tabular}
\caption{
Mean reciprocal rank (MRR) on the \emph{test} set of \data. The best-performing model in each category is \textbf{bold}, and the second best is {\underline{underlined}}. Reasoning types: \textbf{COM}: comparison, \textbf{DE}: data extraction, \textbf{LOC}: location, \textbf{VU}: visual understanding.
}
\label{tab:locality_mrr}

\end{table*}
\begin{table*}[!t]
\centering
\small
\begin{tabular}{@{}lrcccccac@{}}
\toprule
\textbf{Model} 
& \textbf{\begin{tabular}[c]{@{}c@{}}Context \\Window\end{tabular}}
& \textbf{\begin{tabular}[c]{@{}c@{}}CU\end{tabular}} 
& \textbf{\begin{tabular}[c]{@{}c@{}}II\end{tabular}} 
& \textbf{\begin{tabular}[c]{@{}c@{}}RDI\end{tabular}} 
& \textbf{\begin{tabular}[c]{@{}c@{}}MA\end{tabular}} 
& \textbf{\begin{tabular}[c]{@{}c@{}}CA\end{tabular}} 
& \textbf{\begin{tabular}[c]{@{}c@{}}All\end{tabular}} 
\\ \midrule

Expert Performance &  & 72.32 & 71.11 & 83.15 & 76.84 & 79.17 & 76.50\\ 

\midrule
\multicolumn{8}{c}{\textbf{Open-Source Large Language Models (LLMs)}} \\ 
\midrule
$^\dagger$Command R+ & 128,000  & \bb{40.00} & 22.73 & \bb{33.33} & \bb{37.91} & \bb{39.53} & \bb{33.25}\\
Llama-3-70B & 8192 & \underline{31.35} & \bb{35.23} & 22.84 & 32.49 & 35.19 & \underline{31.30} \\
Mistral-7B & 32,768  & 17.10 & 24.09 & 8.89 & 25.81 & 26.72 & 20.45\\
PaLM-2 & 36,864 & 20.73 & 26.42 & 16.35 & 27.65 & 26.72  & 23.55\\
DBRX & 32,768  & 18.13 & 19.43 & 13.94 & 21.30 & 22.63 & 19.05 \\
Gemma-7B & 8,192 & 8.89 & 15.15 & 1.39 & 13.95 & 20.93  & 12.25 \\
\midrule
\multicolumn{8}{c}{\textbf{Proprietary Large Language Models (LLMs)}} \\ 
\midrule
$^\dagger$GPT-3.5 & 16,385 & 22.22 & \underline{33.33} & 19.44 & \underline{32.56} & \underline{37.21} & 29.00 \\
$^\dagger$GPT-4 & 128,000 & 31.11 & 21.21 & \underline{23.61} & \underline{32.56} & 31.40 & 28.50 \\
$^\dagger$Claude-3-Haiku & 200,000 & 28.89 & 30.88 & 12.50 & 29.07 & 38.10  & 28.25  \\ 
$^\dagger$Claude-3-Sonnet & 200,000 & 25.56 & 21.21 & 19.44 & 25.58 & 38.37 & 26.50 \\
$^\dagger$Claude-3-Opus  & 200,000 & 26.67 & 18.18 & 20.83 & 26.74 & 30.23 & 25.00\\
$^\dagger$Gemini-Pro-1.0 & 30,720 & 18.89 & 19.70 & 18.06 & 22.09 & 29.07  & 21.75\\
\bottomrule
\end{tabular}

\caption{
LLM-based accuracy score on the \emph{test} set of \data in \emph{retrieval} setting from GPT-4o's ranking. The best-performing model in each category is \textbf{bold}, and the second best is {\underline{underlined}}. Human expert performance is assessed in an oracle setting, where the correct reference paper is pre-identified. Reasoning types: \textbf{CU}: conceptual understanding, \textbf{II}: implications and Inferences, \textbf{RDI}: results and data interpretation, \textbf{MA}: methodological analysis, \textbf{CA}: critical analysis. $^\dagger$: Due to budget constraints, we randomly sampled 200 instances from the \emph{test} set for evaluation.
}
\label{tab:detail_gpt_based}
\end{table*}

\section{Experiments}

We evaluate 18 foundation models, including both \emph{open-source} and \emph{proprietary} LMMs and LLMs. 
For each model, we select the most recent, largest, and best-performing checkpoint as of April 15th, 2024. 
The evaluation of the \data benchmark is structured into two distinct stages: \emph{visual context} evaluation and \emph{reference-based} evaluation.

\subsection{Visual Context Evaluation}

\paragraph{Task Formulation.}
The visual context evaluation with LMMs is defined as follows: 
Given a visual context question $Q_{vis}$, its correspondent scientific image $I$, and a list of reference papers $D = \{d_1, d_2, \cdots, d_n\}$, the objective is to determine a ranking of these papers based on their relevance to the question and the image. 
This ranking is represented by $R = \{r_1, r_2, \cdots, r_n\}$, where $r_i$ denotes the ranking of the paper $d_i$ for each index $i \in \{1, 2, \cdots, n\}$.
We input $Q_{vis}$, $I$ and $D$ into each LMM, denoted by $f_{LMM}$, and instruct it to generate a ranking $R$ of $D$ based on their relevance to $Q_{vis}$ and $I$:
\begin{equation*}
    R = f_{LMM}(Q_{vis}, I, d_1, d_2, \cdots, d_n)
\end{equation*}
For comparative analysis, simple baselines presented in Table \ref{tab:locality_mrr} are also assessed for the ranking task. 
Other than BM25, these baselines employ cosine similarity between query and document embeddings to rank documents. 
Each query combines the visual context question $Q_{vis}$ and its image caption $C$ generated by GPT-4o with one of the documents, represented by its title and abstract. 
Given a visual context question $Q_{vis}$, its correspondent scientific image $I$, a list of reference papers $D = \{d_1, d_2, \cdots, d_n\}$, an embedding model $\textit{Embed}$, and a cosine similarity function $\textit{sim}$, the ranking process is defined as below:
\begin{equation*}
    \begin{aligned}
        &C = \textit{GPT-4o}(I)\\
        &q = \textit{Embed}(\textit{concat}(Q_{vis}, C))\\
        &\forall d_i \in D, h_i = \textit{Embed}(d_i)\\
        &R = \textit{sort}(sim(q, h_1), \cdots, sim(q, h_n))
    \end{aligned}
\end{equation*}

\paragraph{Evaluation Protocol.}

At the visual context evaluation stage, we assess LMMs' ability to accurately retrieve and rank the correct reference paper from a complete list of reference papers. 
Performance is measured using an established information retrieval metric, Mean Reciprocal Rank (MRR), which effectively gauges a model's ability to identify and prioritize the most relevant reference paper. 
We also calculate Recall@k and nDCG@k to further analyze LMMs' retrieval effectiveness, with results detailed in Table \ref{tab:recall@k} and \ref{tab:ndcg@k} in Appendix \ref{more_result_analysis}.

\paragraph{Experiment Setup.}
This stage involves five \emph{open-source} LMMs, including \emph{open-source} models, such as 
LLaVA 1.6 \cite{liu2023visual}, 
InternVL-Chat-1.1V \cite{chen2024internvl}, 
Yi-VL-34B \citep{ai2024yi},
DeepSeek-VL \citep{deepseekvl},
and Qwen-VL-Plus \cite{bai2023qwenvl}; 
six \emph{proprietary} LMMs, including 
GPT-4V(ision) \cite{openai2024gpt4},
GPT-4o \cite{gpt-4o}, 
Claude 3 Haiku \citep{claude3},
Claude 3 Sonnet \citep{claude3},
Claude 3 Opus \citep{claude3},
and Gemini Vision Pro 1.0 \cite{geminiteam2023gemini}; 
and five \emph{simple baselines}, including 
BM25, 
Contriever \cite{izacard2021contriever}, 
and OpenAI Embeddings\footnote{\url{https://platform.openai.com/docs/guides/embeddings}} (\texttt{Large}, \texttt{Small}, and \texttt{Ada}).

\subsection{Reference-Based Evaluation}

\paragraph{Task Formulation.}
The reference-based evaluation is defined as follows:
Given a combined question $Q_{comb}$ and a ranking $R$ of the reference papers obtained in the \emph{visual context} evaluation stage, the objective is to answer the question based on the top $k$ ranked paper in $R$, denoted by $\textit{Top}_k(R) = \{R[1], R[2], \cdots, R[k]\}$. 
Since \emph{combined} questions contain elements from both \emph{visual context} and \emph{reference-based} questions, we instruct LLMs to solely concentrate on the \emph{reference-based} aspect of $Q_{comb}$. 
The prompts used for this instruction are detailed in Table \ref{tab:detail prompt} in Appendix \ref{appx:prompt_for_combined_questions}.
Accordingly, we input $Q_{comb}$ and $\textit{Top}_k(R)$ into LLMs, denoted by $f_{LLM}$, and instruct LLMs to answer $Q_{comb}$ based on the textual content in top $k$ ranked papers: 
\begin{equation*}
    \textit{Ans} = f_{LLM}(Q_{comb}, R[1], R[2], \cdots, R[k])
\end{equation*}

\paragraph{Evaluation Protocol.}

At the reference-based evaluation stage, we assesses how LLMs perform on reference-based questions using the top three ranked papers identified from the visual context evaluation stage as context. 
Specifically, these papers are ranked by GPT-4o, which is highlighted as the most effective retrieval model in Table \ref{tab:locality_mrr}. GPT-4o achieves an MRR of 0.488, suggesting that the correct reference paper typically appears in the 2.1-th position, placing it within the top three ranked papers on average.
Given that both reference-based question and answer generation utilize plain text extracted from TeX files, we employ the same parsed TeX files as input for LLMs to solve the text-only, reference-based questions.

\paragraph{Generative Response Metrics.}

Following effectiveness of LLMs in evaluating the quality of short AI-generated responses \cite{wang2023evaluating, lu2024mathvista, alpacafarm, charxiv}, we utilize a strong LLM-evaluator (GPT-4) to evaluate the quality of responses generated in the reference-based evaluation stage.
Specifically, the LLM-evaluator rates answers generated against the \emph{gold answers} using a scoring scale of 0, 0.5, and 1.
To more closely align our scoring scale with expert assessments, we compute \emph{Cohen's Kappa} \cite{cohen's_kappa} to assess the agreement between the LLM-evaluator and expert annotators. 
This comparison is conducted for both the 0-0.5-1 and the 1-2-3-4-5 scales, with prompts utilized for evaluation provided in Table \ref{tab:gpt_based_evaluator_app} in Appendix~\ref{cohen's kappa}. 
Expert annotators are tasked with rating 200 responses from four different LLMs (Command R+, GPT-4, Mistral, and Gemma) using both scales. 
Our calculations reveal a Cohen's Kappa value of 0.520 for the 0-0.5-1 scale and 0.444 for the 1-2-3-4-5 scale. 
These results demonstrate greater consistency with expert evaluations when using the 0-0.5-1 scale. 
Further details and comparative results are presented in Appendix~\ref{cohen's kappa}.
Thus, we adopt the 0-0.5-1 scoring scale for our evaluations.
Additionally, we employ established metrics such as ROUGE \cite{lin-2004-rouge}, BERTScore \cite{zhang2020bertscore}, and AutoACU \cite{autoacu} to further gauge the quality of the generated responses.
Detailed results are provided in Table \ref{tab:ROUGE}, \ref{tab:bertscore}, \ref{tab:autoacu} in Appendix \ref{more_result_analysis}.

\paragraph{Experiment Setup.}

This stage involves six \emph{open-source} Text-Only LLMs, including
Mistral-7B \cite{jiang2023mistral}, 
Llama-3-70B \cite{llama3}, 
DBRX \cite{DBRX}, 
PaLM-2 \citep{palm2},
Gemma \citep{gemma},
and Command R+ \cite{commandR+};
and six \emph{proprietary} LLMs, including 
GPT-3.5 \citep{gpt3.5}, 
GPT-4 \cite{openai2024gpt4}, 
Claude 3 Haiku \citep{claude3}, 
Claude 3 Sonnet \citep{claude3}, 
Claude 3 Opus \cite{claude3}, 
and Gemini-Pro-1.0 \cite{geminiteam2023gemini}.

\begin{figure*}[!t]
\centering
\includegraphics[width=1\textwidth]{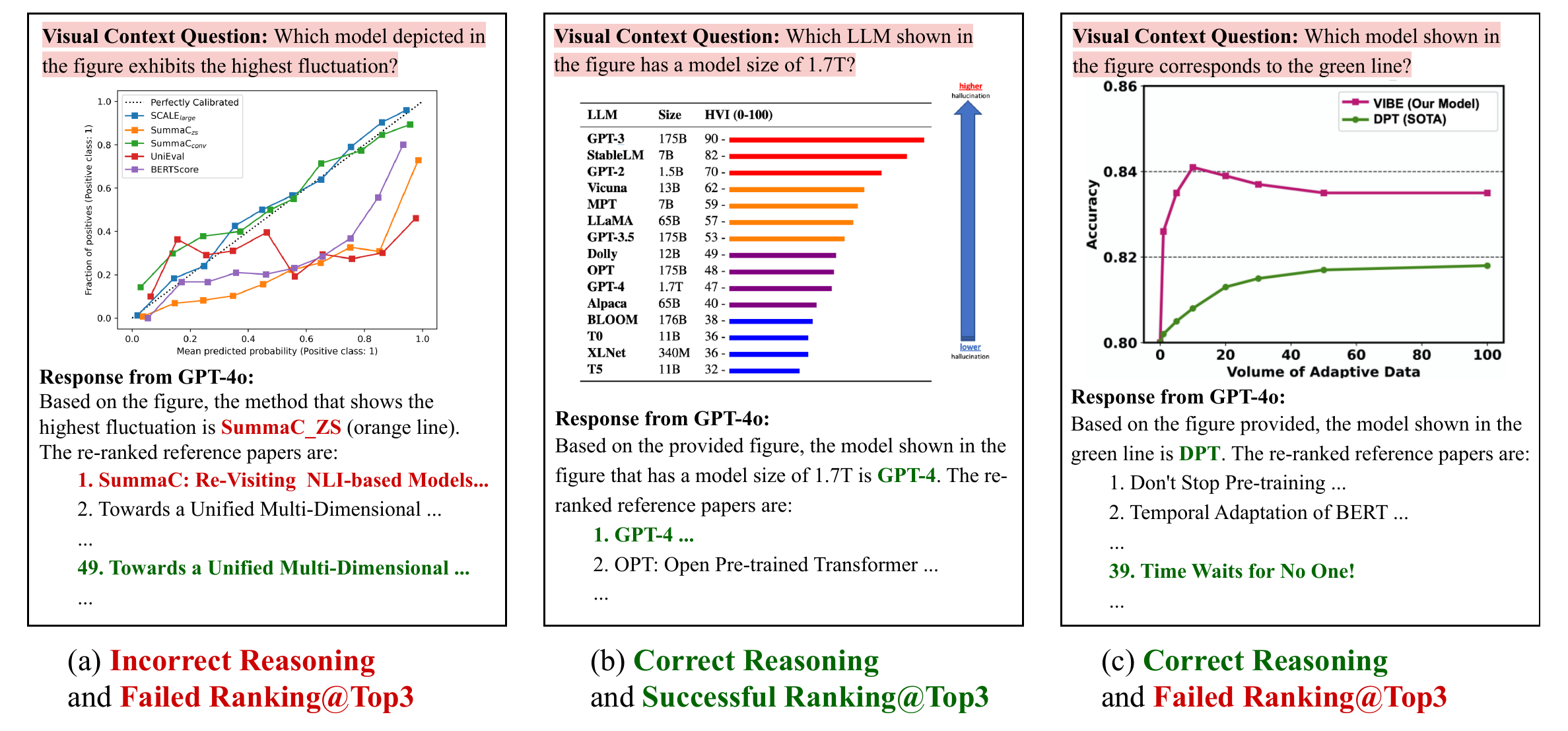}
\caption{Three examples from GPT-4o in answering visual context questions.
} 
\label{fig:fine-grained}
\end{figure*}
\subsection{Main Results}

\label{main_result}
Table \ref{tab:locality_mrr} and Table \ref{tab:detail_gpt_based} present our main results for both \emph{open-source} and \emph{proprietary} LMMs and LLMs on the validation and test set of \data, focusing on \emph{visual context} and \emph{reference-based} questions, respectively. 
We summarize our key findings as follows:

\paragraph{Finding 1: Challenges in Visual Reasoning and Paper Ranking with \data.}
Table \ref{tab:locality_error_analysis} provides a breakdown of GPT-4o's performance in answering the visual context questions, categorized by both reasoning and ranking outcomes. 
Despite being the overall best-performing retriever, GPT-4o still struggles with the dual challenges: it fails to correctly interpret 42.4\% of the scientific images; even when it does produce correct visual reasoning, it falls short in ranking the associated paper within the top three choices. 
Notably, one interesting error pattern is the scenario ``\xmark reasoning \cmark ranking@top3,'' which accounts for 19.7\% of the cases for GPT-4o. 
While this type of error occurs in both open-source and proprietary LMMs, it is more prevalent in the former.
Example error analyses are presented in Figure \ref{fig:fine-grained}, offering a more granular view of these patterns and specific instances where the model underperforms. 

\begin{table}[!t]
  \centering
  \small
  \begin{tabular}{ccc}
    \toprule
    \textbf{\begin{tabular}[c]{@{}c@{}}Reasoning \\Correctness\end{tabular}} & 
    \textbf{\begin{tabular}[c]{@{}c@{}}Ranking@Top3\end{tabular}} & 
    \textbf{\begin{tabular}[c]{@{}c@{}}Percentage\end{tabular}} \\
    \midrule
    \cmark & \cmark & 33.0\% \\
    \cmark & \xmark & 24.7\% \\
    \xmark & \cmark & 19.7\% \\
    \xmark & \xmark & 22.7\% \\
    \bottomrule
  \end{tabular}
  \caption{Performance distribution for GPT-4o on visual context questions, categorized by \bb{Reasoning Correctness} and \bb{Ranking@Top3}.}
  \label{tab:locality_error_analysis}
\end{table}

\paragraph{Finding 2: Inherent Limitations of \emph{Open-Source} LMMs in Long-Range Ranking Task.}
The performance of open-source LMMs in long-range ranking tasks is significantly hindered by their fundamental limitations. We identify three primary challenges:
(1) \emph{Limited Context Window}, which necessitates division of large paper clusters into smaller segments, complicating the ranking process and potentially omitting relevant reference papers;
(2) \emph{Hallucinations}, characterized by the erroneous generation and prioritization of irrelevant arXiv webpage URLs, professional NLP terms, repetitive paper IDs, and random numerical values;
(3) \emph{Formatting Issues}, where models disregard specified format and list papers in plain text, complicating the integration of results across rankings from segmented paper clusters.
These challenges significantly impede the models' ability to provide a comprehensive evaluation of their visual reasoning capabilities, suggesting the need for improvements in their basic functionality to handle more complex reasoning and ranking tasks. A detailed evaluation of open-source LMMs is presented in Appendix \ref{app:LMMs_in_locality}.

\begin{figure}[!hbt]
\centering
\includegraphics[width=1\linewidth]{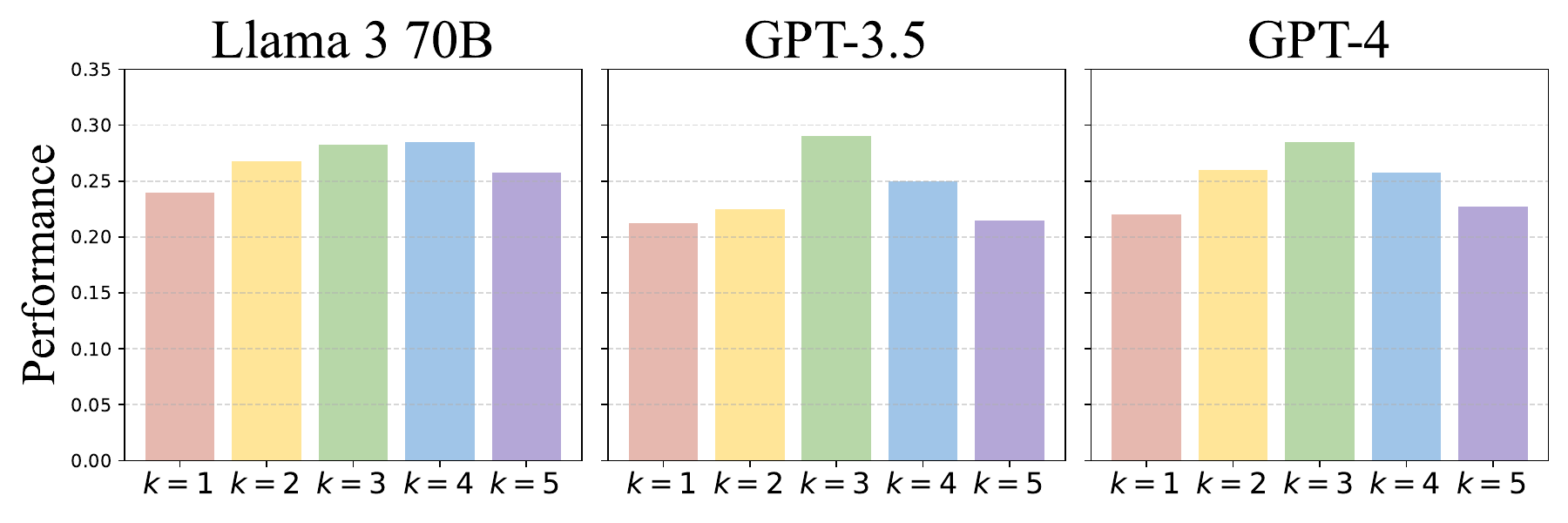}
\caption{Performance scores of Mistral, Llama 3 70B, GPT-3.5, and GPT-4 in different \emph{retrieval} settings.}

\label{fig:performance_score_different_k}
\end{figure}

\paragraph{Finding 3: Precision-Recall Balance.} 
We evaluate LLMs in retrieval settings using the top $k$ ranked papers from the \emph{visual context} evaluation performed by GPT-4o for the values $k \in \{1, 2, 3, 4, 5\}$. 
As shown in Figure \ref{fig:performance_score_different_k}, performance generally increases from $k = 1$ to $k = 3$, aligning with an MRR score of approximately 0.488, which places the correct reference paper in the 2.1-th position on average. Beyond this point, as more papers are considered, more noise is introduced. The general decline in performance after $k = 3$ demonstrates models' limitations in retrieval tasks when given more irrelevant information.

\paragraph{Finding 4: Challenges in Instruction Compliance for LLMs in Retrieval Task.} 
Our evaluation of four models in both a \emph{title-only} setting, where only the title of the reference paper is provided, and a \emph{retrieval} setting, with the top three ranked papers by GPT-4o, highlights variations in instruction compliance. 
Models are instructed to answer ``I don't know'' if a definitive answer cannot be derived from the given information. 
This directive tests the models' adherence to instructions, since the task is infeasible with the titles alone and compliant models should exhibit minimal performance.
Transition to the \emph{retrieval} setting should reveal a significant increase for the models, as observed with GPT-4 in Table \ref{tab:finding4}.
Additionally, employing a LLM-based evaluator to assess generative response overlooks models' confidence levels. Less compliant models, relying on pre-trained knowledge, often produce tangentially relevant responses rather than the instructed ``I don't know,'' leading to disproportionately higher evaluations from the LLM-based evaluator.

\definecolor{deepgreen}{rgb}{0.0, 0.5, 0.0}

\begin{table}[!h]
  \centering
  \small
  \begin{tabular}{lcccc}
    \toprule
    \bb{Models} & GPT-4 & GPT-3.5 & Llama 3 70B & Mistral\\
    \midrule
    \emph{title-only} & 7.50 & 13.50 & 19.75 & 22.00 \\
    \addlinespace[0.2em]\hdashline\addlinespace[0.2em]
    \emph{retrieval} & 
    \begin{tabular}[c]{@{}c@{}}28.50 \\ \textcolor{deepgreen}{\scriptsize (+21.00)}\end{tabular} & 
    \begin{tabular}[c]{@{}c@{}}29.00 \\ \textcolor{deepgreen}{\scriptsize (+15.50)}\end{tabular} &
    \begin{tabular}[c]{@{}c@{}}28.25 \\ \textcolor{deepgreen}{\scriptsize (+8.50)}\end{tabular} & 
    \begin{tabular}[c]{@{}c@{}}19.25 \\ \textcolor{red}{\scriptsize (-2.75)}\end{tabular} \\
    \midrule
  \end{tabular}
  \caption{Performance of four LLMs in answering reference-based questions in \emph{title-only} and \emph{retrieval} setting.}
  \label{tab:finding4}
\end{table}
\section{Related Work}

\paragraph{Multi-Modal QA.} 
Multi-modal QA datasets have posed visual reasoning challenges for LMMs. 
Initially, the focus of benchmarks \cite{coco, MobasherParsVQACapsAB, yagcioglu-etal-2018-recipeqa, talmor2021multimodalqa, scienceqa, chang2022webqa, li2023seedbench, liu2023mmbench, yu2023mmvet} was on conducting QA tasks over simple images, primarily addressing questions such as understanding objects in an image and performing single-hop reasoning. 
Recently, more complex and nuanced benchmarks \cite{chen2022geoqa, lu2024mathvista} have emerged beyond the scope of understanding simple images to require complex mathematical reasoning over diagrams and plots. 
Beyond the scope of mathematical reasoning, MMMU \cite{yue2023mmmu} requires more complex visual reasoning in a diverse range of subjects including science, humanities, and engineering. 

\paragraph{Document QA.} 
Document QA is crucial in the field of NLP, focusing on extracting, synthesizing, and analyzing information from structured and unstructured documents. 
Early document QA benchmarks \cite{rajpurkar2016squad, bajaj2018msmarco, yang2018hotpotqa} involved short document QA, where questions were posed based on content from web pages such as those in Bing's search results or Wikipedia articles. 
Scientific paper QA benchmarks \cite{dasigi-etal-2021-dataset, pmlr-v202-lee23n} require LLMs to conduct multi-hop reasoning and long-context information processing. 
However, a notable gap exists in the integration of Multi-modal QA with Document QA, particularly in the context of scientific research, where it encompasses a blend of textual and visual data alongside complex textual information. 
\data, bridging this gap, is a benchmark for evaluating foundation models' abilities in both multi-modal and multi-document reasoning.

\section{Conclusion}
Existing scientific QA benchmarks often overlook the complexity of real research workflows, which require interpreting non-textual data and aggregating information from multiple documents.
To bridge this gap, we present \data, a novel multi-modal multi-document scientific QA benchmark designed to  evaluate foundation models. 
Our evaluation and analysis underscore the challenges LMMs face in scientific diagram understanding and long-range information ranking tasks, highlighting the limitations of current models in handling complex scientific documents.
We hope this work paves the way for advancements in multi-modal and long-document understanding.
\section*{Limitations}
\label{limitations}
The evaluations presented in this study are met with certain limitations due to inherent disparities in the context window of current \emph{open-source} and \emph{proprietary} LLMs and LMMs. 
There is a significant difference in context window length between models such as GPT-4 Turbo and Claude-3, which can rank all papers in a paper cluster, and models such as InternVL-Chat-V1.1 and QwenVL, which are restricted to handling only two to eight papers in a single prompt. 
This discrepancy may lead to an ``unfair'' comparison of their capabilities. 
Future work could focus on standardizing or extending the context windows in LMMs to mitigate this issue. 

Furthermore, as discussed in Section~\ref{main_result}, prompting an LMM with a set of possible reference papers may be suboptimal due to the challenges models face in ranking a large number of papers. An alternative approach could involve assessing the relevance of each paper individually by encoding the paper into a textual embedding, then comparing it with the textual embedding with of the visual context question combined with the image representation of the figure. This method could potentially alleviate the challenges of requiring an LMM to sift through a large set of possible reference papers and would be an interesting area for future research.

Additionally, our approach to ranking papers for certain models, in particular BM25 and Contriever, involves using GPT-4o's textual descriptions of images rather than its direct image embedding, which might not accurately capture the nuances of scientific images. 
Current image embedding models such as LLaVA \cite{liu2023visual} and CLIP \cite{radford2021learning}, while proficient with natural images, are not trained on scientific images. 
Developing a specialized LMM trained specifically on scientific images~\cite{li2024multimodal, wu2024scimmir} could potentially enhance its performance in interpreting scientific plots, figures, and tables, thereby improving its potential usage in scientific applications. 
\section*{Acknowledgements}
This project was supported in part by Tata Sons Private Limited, Tata Consultancy Services Limited, and Titan.
We are grateful for the compute support provided by Microsoft Research's AFMR program.
We thank Xinyi Han, Zhongjie Wu, and Amy Zhao for their help in initial stages of this project.

\bibliography{anthology, custom}

\appendix
\addtocontents{toc}{\protect\setcounter{tocdepth}{3}}

\hypersetup{linkcolor=black}
\tableofcontents

\section{Data Collection Guidelines}
\label{appendix:data-collection}
\subsection{Visual Context Reasoning Definition}
\label{app:locality_reasoning_def}
Four visual context question reasoning types are defined in Table \ref{tab:locality_reasoning}.

\begin{table*}[!t]
\centering
\small
\renewcommand\tabcolsep{1.0pt} 
\begin{tabular}{cp{11.5cm}}
    \toprule
    \textbf{Visual Context Reasoning} & \textbf{Description} \\
    \midrule
    \multirow{3}{*}{\parbox{4cm}{\centering Comparison \\ }}
    &  It focuses on evaluating and contrasting information presented in tables, figures, or other data formats. To answer questions of this type, one must analyze and compare specific subjects or variables within the given dataset. \\
    \midrule
    \multirow{2}{*}{\parbox{4cm}{\centering Data Extraction  }}
    &  It directly retrieves specific information from a table or figure. This approach focuses on pinpointing exact data points or details.\\
    \midrule
    \multirow{3}{*}{\parbox{4cm}{\centering Location  \\ }}
    &  It is centered on pinpointing spatial or positional information from a table or figure. This involves identifying either relative or absolute locations, such as the placement of items in a figure or row information in a table.\\
    \midrule
    \multirow{3}{*}{\parbox{4cm}{\centering Visual Understanding \\ }}
    & It emphasizes understanding visual information from the figure, such as colors, shapes, and marker types. This approach involves analyzing and extracting visual information.\\
    \bottomrule
\end{tabular}
\caption{Definitions of four visual context reasoning categories in \data.}
\label{tab:locality_reasoning}
\end{table*}

\subsection{Visual Context Reasoning Examples}
\label{app:locality_reasoning_examples}
Four visual context reasoning types examples are shown in Figure \ref{fig:locality_example}.
\begin{figure*}
    \centering
    \includegraphics[width=\textwidth]{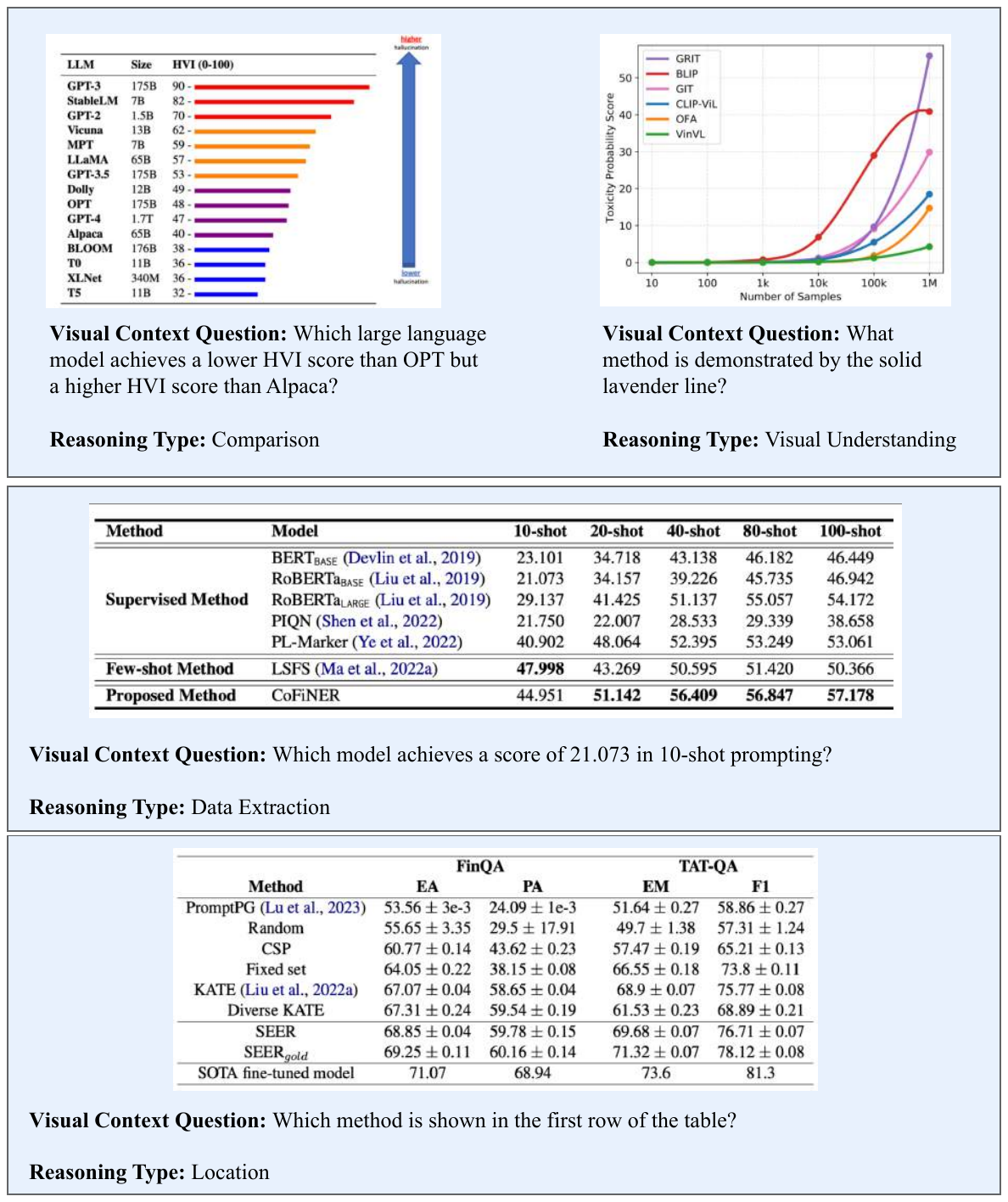}
    \caption{Examples of four visual context reasoning categories in \data.
    }
    \label{fig:locality_example}
\end{figure*}

\subsection{Reference-based Reasoning Definition and Examples}
\label{app:detail_reasoning_def}
Five reference-based question reasoning types and examples are defined in Table \ref{tab:detail_reasoning_definition}.

\begin{table*}[!t]
\centering
\small
\renewcommand\tabcolsep{1.0pt} 
\begin{tabular}{cp{11.5cm}}
    \toprule
    \textbf{Reference-based Reasoning} & \textbf{Description \& Example} \\
    \midrule
    \multirow{3}{*}{\parbox{4cm}{\centering Conceptual Understanding \\ }}
    &  Evaluate knowledge of essential concepts, basic theories, and critical definitions related to the subject. \bb{\emph{Example: What does the hypernetwork in the proposed Hyperdecoders approach generate?}}\\
    \midrule
    \multirow{3}{*}{\parbox{4cm}{\centering Methodological Analysis \\ }}
    & Examine and assess the research methodologies and experimental frameworks employed in studies, with an emphasis on their efficacy and constraints. \bb{\emph{Example: What potential application of the Hyperdecoder approach is suggested by its performance on long-context out-of-domain datasets in the MRQA evaluation?}} \\
    \midrule
    \multirow{3}{*}{\parbox{4cm}{\centering Results and Data Interpretation \\ }}
    & Analyze statistical data, graphs, and tables, focusing on deriving significant insights and conclusions from quantitative and visual information. \bb{\emph{Example: In the experimental results for the GLUE benchmark using $\textbf{T5}_{\textbf{large}}$ v1.1 + LM as the underlying model, which model configuration achieved the highest average score across tasks?}} \\
    \midrule
    \multirow{3}{*}{\parbox{4cm}{\centering Implications and Inferences \\ }}
    & Infer wider implications and practical uses of study outcomes, concentrating on the extensive impact and prospective significance of the results. \bb{\emph{Example: How does the exponentially weighted pooling method in CET ensure that every embedding receives sufficient training?}}\\
    \midrule
    \multirow{3}{*}{\parbox{4cm}{\centering Critical Analysis \\ }}
    & Assess the study’s reasoning, robustness of evidence, and validity of conclusions critically, with a focus on logical consistency and the support of empirical data. \bb{\emph{Example: How does the unified framework's approach to handling the RefCOCOg task diverge in performance between the VL-T5 and VL-BART models?}} \\
    \bottomrule
\end{tabular}
\caption{Definitions of five reasoning categories in \data.}
\label{tab:detail_reasoning_definition}
\end{table*}
\section{Expert Annotation Details}
\label{sec:expert_annotation_details}
\subsection{Expert Annotation for Visual Context Questions}
We employed three computer science graduate students for annotating 300 visual context questions. Being provided with the full list of EMNLP 2023 papers, they were required to: (1) check that each anchor paper has arXiv documentation; (2) find figures or tables that contain comparative information with potential reasoning types described in Table \ref{tab:locality_reasoning}; (3) find the potential reference paper in the figure or table and ensure that it has arXiv documentation; and (4) write the visual context question. When they choose a figure or a table, they were required to fill in the corresponding visual context reasoning type as well as the ``direct answer'' to the visual context question. 

\subsection{Bias Mitigation for Visual Context Questions Annotation}
In preparation for the main annotation process, we conduct a pilot annotation stage where 20 papers where sampled. 
Annotators are instructed to generate three distinct questions per paper.  
These questions are subsequently analyzed by the authors and categorized into four distinct reasoning types: \emph{comparison}, \emph{data extraction}, \emph{location}, \emph{and visual understanding}.
These categories are comprehensive for scientific image understanding. 
By following the predefined reasoning type definitions in Table \ref{tab:locality_reasoning}, we mitigate the risk of annotator bias driven by their own preferences. 
Additionally, these reasoning types are not specific to NLP and are carefully chosen such that they are applicable in analyzing scientific images in the broader scientific fields. 

\subsection{Expert Annotation for Reference-based Questions}
We require each reference paper to have arXiv documentation. Then, we use the arXiv downloader to obtain the full text of the reference paper and generate subsequent reference-based questions (along with answers, explanations, and evidence) using the prompts described in Section \ref{prompt_detail}. We test these questions in the oracle setting, use GPT-based evaluators to evaluate if the answer generated in the oracle setting matches the answer generated along with the question. If they do not match, expert annotators proceeded to manually examine these questions and re-write the answers.

\subsection{Expert Annotation for Reference-based Reasoning}
In Section \ref{prompt_detail}, we automatically assign reasoning types concurrently with the generation of reference-based questions. To ensure the quality of the generated questions, we prompt GPT-4 with the question and its assigned reasoning type to ask if the question matches the reasoning type. For every question that GPT-4 flags as not matching the assigned reasoning type, expert annotators were instructed to manually examine the reasoning types and correct them when necessary. 

\subsection{Expert Annotation for Reference-based Answers}
\label{2_round_checking}

Following the two-round answer generation process mentioned in Section \ref{sec:benchmark_construction}, we manually checked 100 questions for which the first and second round answers matched in order to ensure the gold answers were indeed correct. Out of the 100 sampled questions, 96 questions were marked as correct by expert annotators, demonstrating the high-quality of \data benchmark. 

\section{More Dataset Analysis}

\begin{figure}[p]
    \begin{subfigure}
        \centering
        \hspace{-5mm}
        \includegraphics[width=\columnwidth]{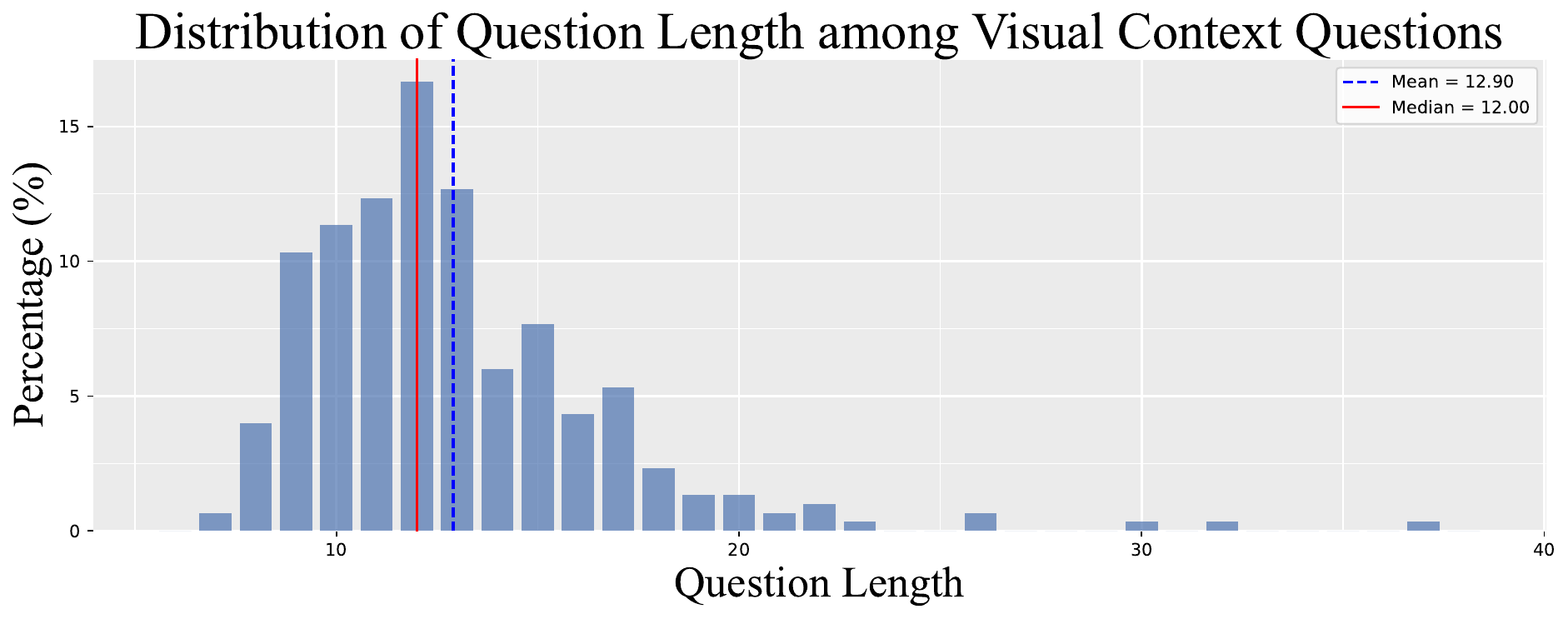}
        \caption{The distribution of the number of tokens per visual context question in \data  - Part 1 of 3.}
        \label{fig:loc_distribution}
    \end{subfigure}
    \begin{subfigure}
        \centering
        \hspace{-5mm}
        \ContinuedFloat
        \includegraphics[width=\columnwidth]{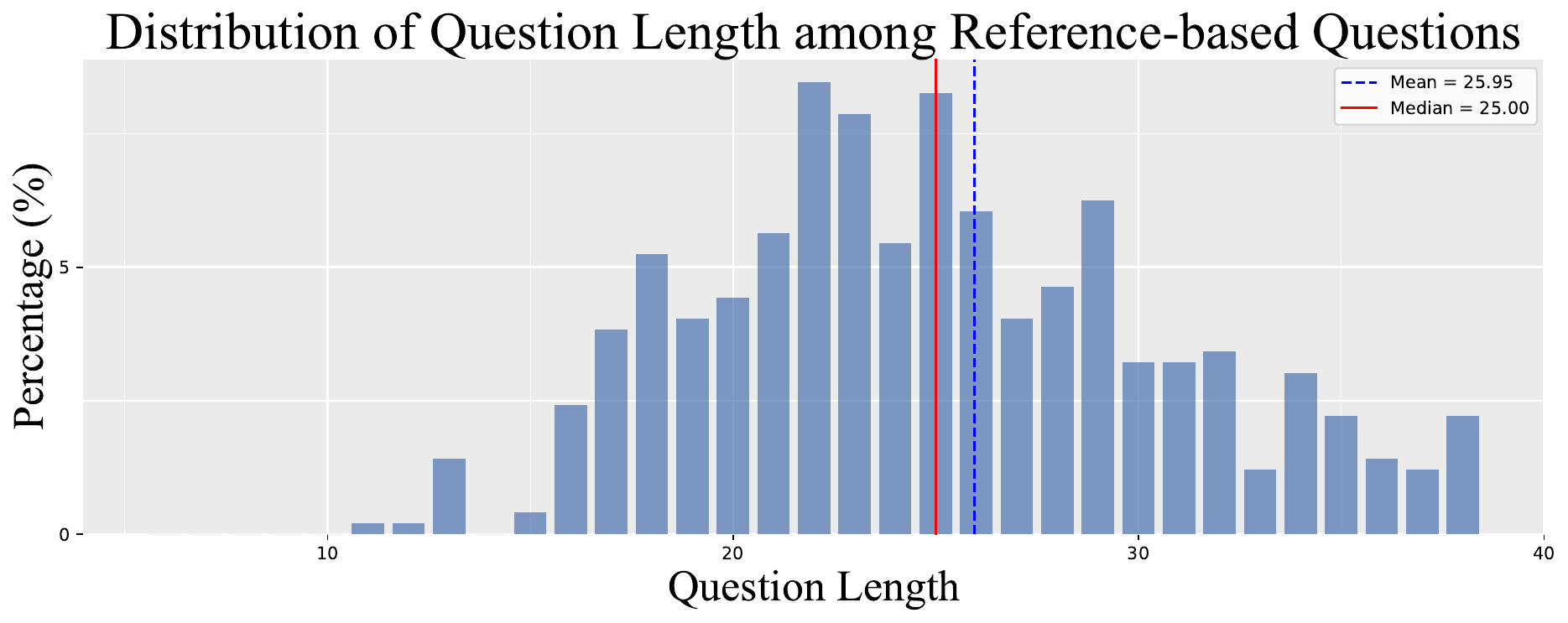}
        \caption{The distribution of the number of tokens per reference-based question in \data  - Part 2 of 3.}
        \label{fig:detail_distribution}
    \end{subfigure}
    \begin{subfigure}
        \centering
        \hspace{-5mm}
        \ContinuedFloat
        \includegraphics[width=\columnwidth]{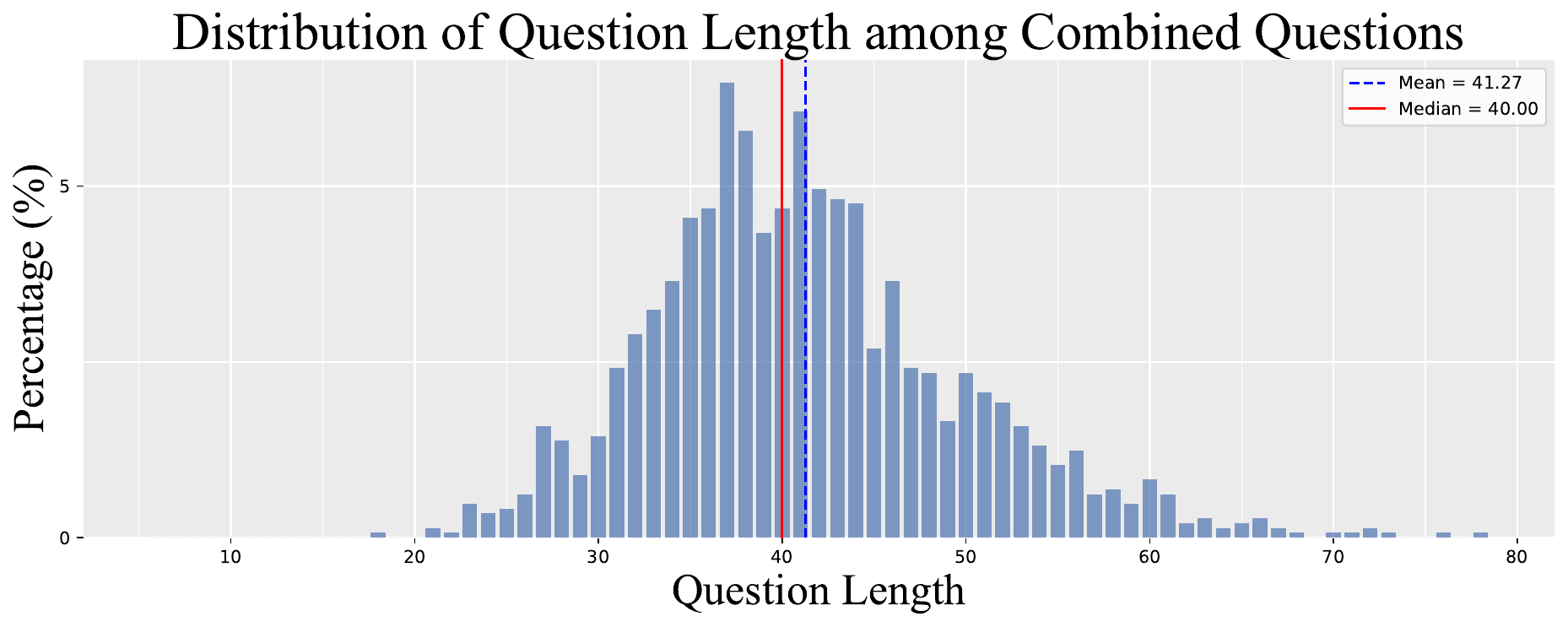}
        \caption{The distribution of the number of tokens per combined question in \data  - Part 3 of 3.}
        \label{fig:combine_distribution}
    \end{subfigure}

    \label{fig:overall_distribution}
\end{figure}

\paragraph{Question Distribution.} As illustrated in Table \ref{tab:dataset_statistics}, the average question length in \data is 41.27 (in tokens), while the maximum number of tokens in a question is 78 (in tokens).

Figure \ref{fig:combine_distribution} further illustrates the distribution of token counts in all visual context, reference-based, and combined questions, highlighting the diverse distribution of all three types of questions. In these figures, the red solid line represents the median and the blue dashed line represents the mean. From all three distributions, we note that the median and mean are very close in values, implying our dataset is symmetric or only slightly skewed.

\section{More Result Analysis}
\label{more_result_analysis}

\paragraph{Recall@k for Visual Context Evaluation.} In addition to the MRR values shown in Table \ref{tab:locality_mrr}, Recall@k is illustrated in Table \ref{tab:recall@k}.

\begin{table}[!h]
\centering
\small
\begin{tabular}{@{}lccccc@{}}
\toprule
\bb{Model} 
& \textbf{\begin{tabular}[c]{@{}c@{}}Recall\\@1\end{tabular} } 
& \textbf{\begin{tabular}[c]{@{}c@{}}Recall\\@3\end{tabular}} 
& \textbf{\begin{tabular}[c]{@{}c@{}}Recall\\@5\end{tabular}} 
\\ 
\midrule
GPT-4o & 0.40 & 0.53 & 0.57\\
GPT-4V(ision) & 0.30 & 0.45 & 0.51 \\
Claude-3-Opus & 0.20 & 0.33 & 0.44\\
Claude-3-Sonnet & 0.30 & 0.46 & 0.57 \\
Claude-3-Haiku & 0.09 & 0.25 & 0.29\\
Gemini-Pro-Vision-1.0 & 0.12 & 0.21 & 0.26\\
\bottomrule
\end{tabular}%

\caption{Recall@k}
\label{tab:recall@k}
\end{table}

\paragraph{nDCG@k for Visual Context Evaluation.} In addition to the MRR values shown in Table \ref{tab:locality_mrr}, nDCG@k is illustrated in Table \ref{tab:ndcg@k}.

\begin{table}[!h]
\centering
\small
\begin{tabular}{@{}lccccc@{}}
\toprule
\bb{Model} 
& \textbf{\begin{tabular}[c]{@{}c@{}}nDCG\\@1\end{tabular} } 
& \textbf{\begin{tabular}[c]{@{}c@{}}nDCG\\@3\end{tabular}} 
& \textbf{\begin{tabular}[c]{@{}c@{}}nDCG\\@5\end{tabular}} 
\\ 
\midrule
GPT-4o & 0.40 & 0.51 & 0.53\\
GPT-4V(ision) & 0.30 & 0.43 & 0.45 \\
Claude-3-Opus & 0.20 & 0.31 & 0.36\\
Claude-3-Sonnet & 0.30 & 0.44 & 0.49 \\
Claude-3-Haiku & 0.09 & 0.23 & 0.25\\
Gemini-Pro-Vision-1.0 & 0.12 & 0.19 & 0.21\\
\bottomrule
\end{tabular}%

\caption{nDCG@k}
\label{tab:ndcg@k}
\end{table}

\paragraph{Standard Metrics for Reference-based Evaluation.} In addition to the LLM-based accuracy results shown in Table \ref{tab:detail_gpt_based}, ROUGE scores are illustrated in Table \ref{tab:ROUGE}; AutoACU scores \cite{autoacu} are illustrated in Table \ref{tab:autoacu}; and each BERTScore \cite{zhang2020bertscore} is provided in Table \ref{tab:bertscore}.

\begin{table}[!h]
\centering
\small
\begin{tabular}{@{}lccc@{}}
\toprule
\bb{Model} 
& \textbf{\begin{tabular}[c]{@{}c@{}}ROUGE-1\end{tabular} } 
& \textbf{\begin{tabular}[c]{@{}c@{}}ROUGE-2\end{tabular}} 
& \textbf{\begin{tabular}[c]{@{}c@{}}ROUGE-l\end{tabular}} 
\\ \midrule
Llama-2-70B & 0.125 & 0.056 & 0.098\\
Mistral-7B & 0.182 & 0.086 & 0.143\\
PaLM-2 & 0.197 & 0.095 & 0.157\\
Gemma-7B & 0.073 & 0.032 & 0.058\\
DBRX & 0.155 & 0.075 & 0.122\\
$^\dagger$Command R+ & 0.205 & 0.079 & \underline{0.176}\\
\midrule
$^\dagger$GPT-4 & \bb{0.237} & \bb{0.127} & \bb{0.202}\\
$^\dagger$GPT-3.5 & \underline{0.208} & 0.101 & 0.171\\
$^\dagger$Gemini-Pro-1.0 & 0.192 & \underline{0.104} & 0.162\\
$^\dagger$Claude-3-Haiku & 0.176 & 0.090 & 0.143\\
$^\dagger$Claude-3-Sonnet & 0.184 & 0.086 & 0.144\\
$^\dagger$Claude-3-Opus & 0.182 & 0.087 & 0.140\\
\bottomrule
\end{tabular}%

\caption{ROUGE score on \emph{test} set of \data in \emph{retrieval} setting from GPT-4V(ision)'s retrieval. The best-performing model in each category is \textbf{bold}, and the second best is {\underline{underlined}}.}
\label{tab:ROUGE}
\end{table}
\begin{table}[!ht]
\centering
\small
\begin{tabular}{@{}lccc@{}}
\toprule
\bb{Model} 
& \textbf{\begin{tabular}[c]{@{}c@{}}Recall\end{tabular} } 
& \textbf{\begin{tabular}[c]{@{}c@{}}Precision\end{tabular}} 
& \textbf{\begin{tabular}[c]{@{}c@{}}F-1\end{tabular}} 
\\ \midrule
Llama-2-70B & 0.852 & 0.807 & 0.828\\
Mistral-7B & 0.855 & 0.832 & 0.843\\
PaLM-2 & 0.855 & 0.843 & 0.848\\
Gemma-7B & 0.359 & 0.355 & 0.357\\
DBRX & 0.721 & 0.698 & 0.709\\
$^\dagger$Command R+ & 0.856 & \bb{0.862} & \bb{0.859}\\
\midrule
$^\dagger$GPT-4 & \bb{0.865} & \underline{0.851} & \underline{0.858}\\
$^\dagger$GPT-3.5 & \underline{0.861} & 0.842 & 0.851\\
$^\dagger$Gemini-Pro-1.0 & 0.852 & 0.847 & 0.849\\
$^\dagger$Claude-3-Haiku & 0.855 & 0.827 & 0.840\\
$^\dagger$Claude-3-Sonnet & 0.856 & 0.829 & 0.842\\
$^\dagger$Claude-3-Opus & 0.855 & 0.827 & 0.840\\
\bottomrule
\end{tabular}%

\caption{BERTScore on \emph{test} set of \data in \emph{retrieval} setting from GPT-4V(ision)'s retrieval. The best-performing model in each category is \textbf{bold}, and the second best is {\underline{underlined}}.}
\label{tab:bertscore}
\end{table}
\begin{table}[!h]
\centering
\small
\begin{tabular}{@{}lccc@{}}
\toprule
\bb{Model}  & \bb{Recall} & \bb{Precision} & \bb{F-1}\\
\midrule
Llama-2-70B & 0.212 & 0.091 & 0.111 \\
Mistral-7B &0.176 &0.104 &0.109 \\
PaLM-2 & 0.170 &0.123 & 0.113  \\
Gemma-7B & 0.097 &\bb{0.198} &  0.107  \\
DBRX & 0.164 & 0.131&0.111    \\
$^\dagger$Command R+ & 0.155 & 0.153& 0.115   \\
\midrule
$^\dagger$GPT-4 &\bb{0.226} & \underline{0.164}&\bb{0.158}  \\
$^\dagger$GPT-3.5 &0.195 &0.124 &0.118 \\
$^\dagger$Gemini-Pro-1.0 & 0.170 & 0.134& \underline{0.123}  \\
$^\dagger$Claude-3-Haiku & 0.217& 0.113& 0.118 \\
$^\dagger$Claude-3-Sonnet &0.215 & 0.010& 0.110  \\
$^\dagger$Claude-3-Opus & \underline{0.224}& 0.108& 0.116   \\
\bottomrule
\end{tabular}%

\caption{AutoACU (A3CU) score on \emph{test} set of \data in \emph{retrieval} setting from GPT-4V(ision)'s retrieval. The best-performing model in each category is \textbf{bold}, and the second best is {\underline{underlined}}.}
\label{tab:autoacu}
\end{table}

\section{More Details On the Setup}

\subsection{LLM-Based Evaluator.} 
\label{cohen's kappa}
Cohen’s Kappa results are detailed in Table \ref{tab:cohen's kappa}, illustrating the level of concordance between expert annotators and LLM-evaluators. Our result reveals a Cohen's Kappa value of 0.520 for the 0-0.5-1 scale and 0.444 for the 1-2-3-4-5 scale. While the Cohen's Kappa value of 0.520 only indicates a ``weak agreement'' with humans \cite{cohen's_kappa}, the 0-0.5-1 scale demonstrates stronger agreement compared to the 1-2-3-4-5 scale. Additionally, the evaluation prompts used for both the 0-0.5-1 and 1-2-3-4-5 scales are provided in Table \ref{tab:gpt_based_evaluator_app}. 

\begin{table}[!t]
    \centering
    \small
    \begin{tabular}{ccc}
    \toprule
     & \textbf{0-0.5-1} & \textbf{1-2-3-4-5} \\
    \midrule
    Expert Annotators & 0.520 & 0.444 \\
    \bottomrule
    \end{tabular}
    \caption{Cohen's Kappa between two LLM-evaluators \wrt expert annotations.}
    \label{tab:cohen's kappa}
\end{table}

\subsection{Prompt for Evaluating Visual Context Question}
Prompts used to rank reference papers across multiple LMMs are illustrated in Table \ref{tab:locality prompt}.
\begin{table*}[!t]
\centering
\small
\renewcommand\tabcolsep{1.0pt} 
\begin{tabular}{cp{11.5cm}}
    \toprule
    \textbf{Model} & \textbf{Prompt} \\
    \midrule
    \multirow{3}{*}{\parbox{4cm}{\centering Yi-VL-34B\\DeepSeek-VL}}
    &  Answer the question from the figure and the reference papers provided only: \{question\}
    
    Additionally, rerank the following reference papers according to their relevance to this question. Each reference paper consists of an S2\_id, a title, and an abstract.
    
    \{paper\_cluster\}
    
    Format your answer as a python dictionary with keys "question", "answer", and "rank". "rank" should be a list of S2\_id. 
    
    If no relevant reference papers are provided, return an empty list for "rank". Note: The "rank" list should only include **question-relevant** reference papers. Do not include irrelevant ones.\\
    \midrule
    \multirow{3}{*}{\parbox{4cm}{\centering InternVL-Chat-1.1V}}
    &  You are given a figure, a question, and some paper candidates of titles and abstracts. Your task is to answer the question based on the figure information, then order the paper candidates that I provide to you so that the paper that is more relevant to the question comes first in the list.
    Provide your answer at the end in a json file of this format using S2\_id only:\{"ranking":  [""] \}. Make sure the responded list is in a valid format and that it only contains the S2\_id. Do not include the title or abstract in the answer list. 

    <question>
    \{question\}
    </question>

    <paper candidates>
    \{paper\_cluster\}
    </paper candidates>\\
    \midrule
    \multirow{3}{*}{\parbox{4cm}{\centering LLaVa-1.6\\Qwen-VL}}
    & Answer the question from the figure and the reference papers provided only: \{question\}
    
    Additionally, rerank the following reference papers according to their relevance to this question. Each reference paper consists of an S2\_id, a title, and an abstract.
    
    \{paper\_cluster\}
    
    Format your answer as a python dictionary with keys "question", "answer", and "rank". "rank" should be a list of S2\_id. 
    
    If no relevant reference papers are provided, return an empty list for "rank".
    \\
    \midrule
    \multirow{8}{*}{\parbox{4cm}{\centering GPT-4o\\GPT-4V(ision)\\Gemini-Pro-Vision-1.0\\Claude-3-Haiku\\Claude-3-Sonnet\\Claude-3-Opus}}
    & You are given a figure, a question, and a list of paper candidates of titles and abstracts. Your task is to answer the question based on the figure information and then re-rank the list of paper candidates I provided to you. 

        Provide your answer at the end in a json format using the S2\_id only: \{"ranking":  []\}.
        Only include papers that are relevant. Do not include papers that are irrelevant. Make sure the answer list is properly formatted.

        <question>
        \{question\}
        </question>

        <paper candidates>
        \{paper\_cluster\}
        </paper candidates>
    \\
    \bottomrule
\end{tabular}
\caption{Prompts used to rank reference papers across multiple LMMs.}
\label{tab:locality prompt}
\end{table*}

\subsection{Prompt for Answering Reference-based Question}
\label{appx:prompt_for_combined_questions}
Prompts used to answer reference-based questions are illustrated in Table \ref{tab:detail prompt}.

\begin{table*}[!t]
\centering
\small
\renewcommand\tabcolsep{1.0pt} 
\begin{tabular}{cp{11.5cm}}
    \toprule
    \textbf{Stage} & \textbf{Prompt} \\
    \midrule
    \multirow{3}{*}{\parbox{4cm}{\centering Answers from text chunk}}
    &  Answer the below question about a scientific paper. The question is composed of 2 parts, and the second part of the question can be answered from the paper. I will provide you with only a chunk of a paper. 
        
Explain your reasoning. Append the answer at the end of the response in a json format \{``answer'': ``''\}. You should answer the question in a short-answer form. Do not provide long answers. If you do not know the answer, respond with \{``answer'': ``I don't know''\}
        
<QUESTION>
\{question\}
</QUESTION>

<CHUNK>
\{chunk\}
</CHUNK>\\
    \midrule
    \multirow{3}{*}{\parbox{4cm}{\centering Answer aggregation}}
    &  I will provide you with a set of answer candidates for a question. Aggregate the information from all the candidates and give me one single answer. Note that if one answer candidate is `I don't know', you can ignore it.
    Answer the question based on the answer candidates and summarize the final answer into a short answer.
    <QUESTION>
    \{question\}
    </QUESTION>
    \{answer\_candidate\_list\}\\

    \bottomrule
\end{tabular}
\caption{Prompts used to generate and aggregate answers from a text chunk.}
\label{tab:detail prompt}
\end{table*}

\subsection{Prompt for Answer Evaluation}
Prompts used to retrieve answers from each text chunk and aggregate answers are illustrated in Table \ref{tab:gpt_based_evaluator_app}.
\begin{table*}[!t]
\centering
\small
\renewcommand\tabcolsep{1.0pt} 
\begin{tabular}{cp{11.5cm}}
    \toprule
    \textbf{Evaluator} & \textbf{Prompt} \\
    \midrule
    \multirow{8}{*}{\parbox{4cm}{\centering LLM-based Evaluator \\(0-0.5-1 setting)}}
    &  I am testing a model's performance on open-ended questions. I want you to help me in checking to see if the candidate answer has the same meaning as the reference answer for a given question. If you think the reference answer and the candidate answer have the same meaning, respond \{``selection'': ``1''\}; otherwise, respond by \{``selection'': ``0''\}. If you think the candidate is partially correct, respond by \{``selection'': ``0.5''\}. If the answer is ``I don't know,'' rate it to 0.

    <QUESTION>
    \{question\}
    </QUESTION>

    <REFERENCE>
    \{reference\}
    </REFERENCE>

    <CANDIDATE>
    \{candidate\}
    </CANDIDATE>
\\
\midrule

\multirow{28}{*}{\parbox{4cm}{\centering LLM-based Evaluator \\(1-2-3-4-5 setting)}}
    &  I am testing a model's performance on open-ended questions. I want you to help me in checking to see if the candidate answer has the same meaning as the reference answer for a given question. 
    
    Rate the candidate answer from 1, 2, 3, 4, and 5, where 1 means the candidate is the least similar to the reference answer and 5 means the candidate matches to the reference answer perfectly. 
    Respond by \{``selection'': ``''\}. If the candidate answer is ``I don't know,'' rate it to 1.

    Here's some examples you can consider:
    
    Question: Why transformer is better than RNN?
    
    Reference: Parallel computation
    
    Candidate: Computation
    
    Rating: 3

    Question: What's the major advantage of using ALiBi positional embedding?
    
    Reference: Effectively handle sequences of varying lengths, particularly beneficial for very long sequences
    
    Candidate: It has more freedom to handle input
    
    Rating: 2

    Question: What's the model's performance on GSK8K dataset? 
    
    Refernece: 65.65\%
    
    Candidate: 44.56\%
    
    Rating: 1

    Question: What specific method does this paper propose to solve LLM searching problem?
    
    Reference: MCTS
    
    Candiate: Monte Carlo Tree Search is proposed in this paper to solve searching when using decomposed prompting method.
    
    Rating: 5

    Question: How does the performance change when we switch from CoT to ToT in prompting? 
    
    Reference: Accuracy from 23.50\% to 32.87\%
    
    Candidate: slightly increase
    
    Rating: 4

    <QUESTION>
    \{question\}
    </QUESTION>

    <REFERENCE>
    \{reference\}
    </REFERENCE>

    <CANDIDATE>
    \{candidate\}
    </CANDIDATE>
\\
    \bottomrule
\end{tabular}
\caption{Prompts used to evaluate answers generated by LLMs.}
\label{tab:gpt_based_evaluator_app}
\end{table*}

\subsection{Prompt for Reference-based Question Generation}
\label{prompt_detail}
We list our prompt for reference-based question generation in Table \ref{tab:E.5}.

\begin{table*}[!t]
\centering
\small
\renewcommand\tabcolsep{1.0pt} 
\begin{tabular}{cp{11.5cm}}
    \toprule
    \textbf{Model} & \textbf{Prompt} \\
    \midrule

    \multirow{18}{*}{\parbox{4cm}{\centering Reference-based Question Generation Prompt}}
    &  Generate 1 short answer question based on the paper's full content below. You should follow the reasoning type of \{reasoning\_type\}, with the definition \{reasoning\_description\}. The short answer question should be as hard as possible, and focus on a single detail from the paper. The target audience of the short answer question is an expert in the field of natural language processing. The question should be hard for GPT-4 to answer. The answer to the question should be short and must be answerable from the content of the paper.

Here are some requirements: 

<REQUIREMENTS>
[Question] should make sense and can be answered from the paper's full text.
[Answer] should be directly answering the question you generated.
[Explanation] should explain why the answer correctly answers the question.
[Evidence] should be from the original content from the paper content. This should be an excerpt from the input paper that supports your answer. 

</REQUIREMENTS>

Append the answer at the end of your response in a json-like format:

\{``question'': ``'', 

``answer'': ``'', 

``explanation'': ``'', 

``evidence'':``''\}

<PAPER FULL CONTENT>

{full\_text}

</PAPER FULL CONTENT>\\

    \bottomrule
\end{tabular}
\caption{Prompt for reference-based question generation.}
\label{tab:E.5}
\end{table*}

\subsection{Model Parameters for Answering Visual Context Question}
Model parameters for ranking reference papers from a paper cluster are shown in Table \ref{tab:model_param_locality}.
\begin{table*}[!t]
\centering
\small
\begin{tabular}{cl}
    \toprule
    \textbf{Model} & \textbf{Generation Setup} \\
    \midrule
    LLaVa-1.6 & model = \texttt{llava-v1.6-mistral-7b}, temperature = 0.1, max\_tokens = 8192\\
    \midrule
    Yi-VL-6B & model = \texttt{Yi-VL-6B}, temperature = 0.1, max\_tokens = 8192\\
    \midrule
    DeepSeek-VL & model = \texttt{deepseek-vl-7b-chat}, temperature = 0.1, max\_tokens = 8192\\
    \midrule
    InternVL-Chat-1.1V & model = \texttt{InternVL-Chat-Chinese-V1-1}, temperature = 0.1, max\_tokens = 768 \\
    \midrule
    Qwen-VL-Plus & model = \texttt{qwen-vl-plus}, seed = 1234, max\_tokens = 6000\\
    \midrule
    GPT-4o 
    & model = \texttt{gpt-4o}, temperature = 0.1, max\_tokens = 4096\\
    \midrule
    GPT-4V(ision) 
    & model = \texttt{gpt-4-turbo}, temperature = 0.1, max\_tokens = 4096\\
    \midrule
    Gemini-Pro-Vision-1.0 
    & model = \texttt{gemini-pro-vision}, temperature = 0.1, max\_tokens = 4096\\
    \midrule
    Claude-3-Haiku 
    & model = \texttt{claude-3-haiku-20240307}, temperature = 0.1, max\_tokens = 4096\\
    \midrule
    Claude-3-Sonnet 
    & model = \texttt{claude-3-sonnet-20240229}, temperature = 0.1, max\_tokens = 4096\\
    \midrule
    Claude-3-Opus 
    & model = \texttt{claude-3-opus-20240229}, temperature = 0.1, max\_tokens = 4096\\
    \bottomrule
\end{tabular}
\caption{Parameters of various LMMs in evaluating visual context questions.}
\label{tab:model_param_locality}
\end{table*}

\subsection{Model Parameters for Answering Reference-based Question}
Model parameters for answering reference-based questions are exhibited in Table \ref{tab:model_param_detail}.
\begin{table*}[!t]
\centering
\small
\begin{tabular}{cl}
    \toprule
    \textbf{Model} & \textbf{Generation Setup} \\
    \midrule
    Llama-3-70B & temperature = 0.1, max\_token = 10,000  \\
    \midrule
    Mistral-7B & temperature = 0.1, max\_token = 40,000\\
    \midrule
    PaLM-2 & temperature = 0.1, max\_token = 40,000 \\
    \midrule
    Gemma & temperature = 0.1, max\_token = 12,000\\
    \midrule
    DBRX & temperature = 0.1, max\_token = 40,000 \\
    \midrule
    Command R+ & temperature = 0.1, max\_token = 200,000 \\
    \midrule
    GPT-4 
    & model = \texttt{gpt-4-0125-preview}, temperature = 0.1, max\_tokens = 200,000\\
    \midrule
    Gemini-Pro-1.0 
    & model = \texttt{gemini-1.0-pro}, temperature = 0.1, max\_tokens = 40,000\\
    \midrule
    Claude-3-Haiku 
    & model = \texttt{claude-3-haiku-20240307}, temperature = 0.1, max\_tokens = 250,000\\
    \midrule
    Claude-3-Sonnet 
    & model = \texttt{claude-3-sonnet-20240229}, temperature = 0.1, max\_tokens = 250,000\\
    \midrule
    Claude-3-Opus 
    & model = \texttt{claude-3-opus-20240229}, temperature = 0.1, max\_tokens = 250,000\\
    \bottomrule
\end{tabular}
\caption{Parameters of various LLMs in evaluating reference-based questions.}\label{tab:model_param_detail}
\end{table*}

\section{A Comparative Study of LMMs in Answering Visual Context Questions}
\label{app:LMMs_in_locality}

In our experiments, we evaluated numerous LMMs in answering visual context questions, such as Kosmos2, Fuyu-8B, and Qwen-VL-Chat. Our findings indicate that these models severely suffer from both hallucination and formatting errors when analyzing the scientific figures. Thus, we conclude that they lack the basic capabilities to generate valid rankings, which are crucial for calculating MRR.

\subsection{InternVL-Chat-1.1V}

\begin{table}[!t]
  \centering
  \small
  \begin{tabular}{lcc}
    \toprule
    \bb{Models} & \textbf{validation} & \textbf{test} \\
    \midrule
    Method 1 & 0.07 & 0.07 \\
    Method 2 & 0.218 & 0.186 \\
    Method 3 & 0.152 & 0.193 \\
    \bottomrule
  \end{tabular}
  \caption{MRR for InternVL}
  \label{tab:internvl_graular_result}
\end{table}

InternVL-Chat-1.1V operates with a short context window, a restriction that makes answering visual context questions particularly difficult. Although pairwise paper rankings were still possible within the token length restrictions, prompting the model with the entire list of possible reference paper titles and abstracts was not possible. Since the vanilla singular prompting method used to test other models with larger context windows (e.g. GPT-4V) on the visual context question dataset could not be applied to InternVL-Chat-1.1V, we used a slightly different prompting scheme. 

Three different ranking settings and methodologies were used to determine the rank of the reference paper for each visual context question. In the first setting, the model was repeatedly prompted to compare the true reference paper against each of the other papers one at a time in a head-to-head ranking. In this setting, we then considered the true reference paper's rank to be one more than the number of papers individually ranked higher than the true reference paper when compared side-by-side. In the second setting, the model was prompted to assign a rating to each of the sampled reference papers; the ratings were then sorted to generate a final ranking among the papers. Finally, in the third setting, the model randomly paired papers together, with each of the higher ranked papers in each pair considered to be ranked higher than every lower ranked papers. By then iteratively pairing papers among the set of higher-ranked papers and also iteratively pairing papers among all the initially lower-ranked ones, a ranking for the true reference paper was generated. 

Comparing each pair of sample papers requires a quadratic number of queries to the model, which requires a significant amount of time. Each of the three proposed methods, on the other hand, require a number of model queries that is linear in the total number of sample references. 

However, each of the methodologies have their own potential flaws. The first ranking methodology was asymmetric in that the true reference paper was prompted a different number of times; thus, for a method with no reasoning or retrieval capabilities, the true reference paper would have a $1/2^{n-1}$ chance of being ranked first, while it would have a $1/n$ chance of being ranked first in the ranking mechanism used in larger models, if there are $n$ papers to rank. Since MRR heavily favors smaller ranks, the first ranking methodology would bias the observed MRR downward. The second methodology, with zero-shot prompting, was unstable at times; furthermore, the model generally only chose from a set of a few possible ratings (i.e. $0$, $80$, $90$, or $100$ out of $100$), making it hard to differentiate and rank papers with the same rating. The third method is symmetric in its prompting but yields different results depending on initial pairings; we randomize the papers when pairing, and so this method is unbiased. We report the MRR values from the third method in Table \ref{tab:locality_mrr}. Detailed results are illustrated in Table \ref{tab:internvl_graular_result}.

\subsection{Qwen-VL-Plus}
\label{appx:qwen-analysis}
In the visual context evaluation stage, only 53.1\% of Qwen-VL-Plus's rankings are valid, with a mere 5.0\% including the ground truth paper. 
MRR for QwenVL-Plus is evaluated based on 3 criteria: 
(1) \emph{Rank All}, assigning a zero value to any invalid rankings; 
(2) \emph{Rank Valid}, considering only valid rankings for MRR computation; 
and (3) \emph{Rank Ground Truth}, calculating MRR solely from rankings that include the ground truth. 
Detailed findings are presented in Table \ref{tab:qwenvl_breakdown}, though only \emph{Rank Valid} is reported in Table \ref{tab:detail_gpt_based}. Additional Error analysis can refer to Figure \ref{fig:qwen_locality1} below.

\begin{table}[!t]
    \centering
    \small
    \begin{tabular}{cccc}
    \toprule
    \textbf{model} & 
    \textbf{\begin{tabular}[c]{@{}c@{}}Rank \\ All\end{tabular}} & 
    \textbf{\begin{tabular}[c]{@{}c@{}}Rank \\ Valid\end{tabular}} & 
    \textbf{\begin{tabular}[c]{@{}c@{}}Rank \\ Ground Truth\end{tabular}}  \\
    (percentage) & & ($53.1\%$) & ($5.0\%$)\\
    \midrule
    QwenVL-Plus & 0.047 & 0.089 & 0.947 \\
    \bottomrule
    \end{tabular}
    \caption{MRR for QwenVL-Plus on the \emph{test} set across 3 evaluation settings.}
    \label{tab:qwenvl_breakdown}
\end{table}

\subsection{GPT-4V(ision)}
See Figure \ref{fig:gpt_locality1} and Figure \ref{fig:gpt_locality2} below.

\subsection{Claude-3-Opus}
See Figure \ref{fig:opus_locality1} and Figure \ref{fig:opus_locality2} below.

\clearpage
\begin{figure*}[p]
    \centering
    \includegraphics[width=\textwidth]{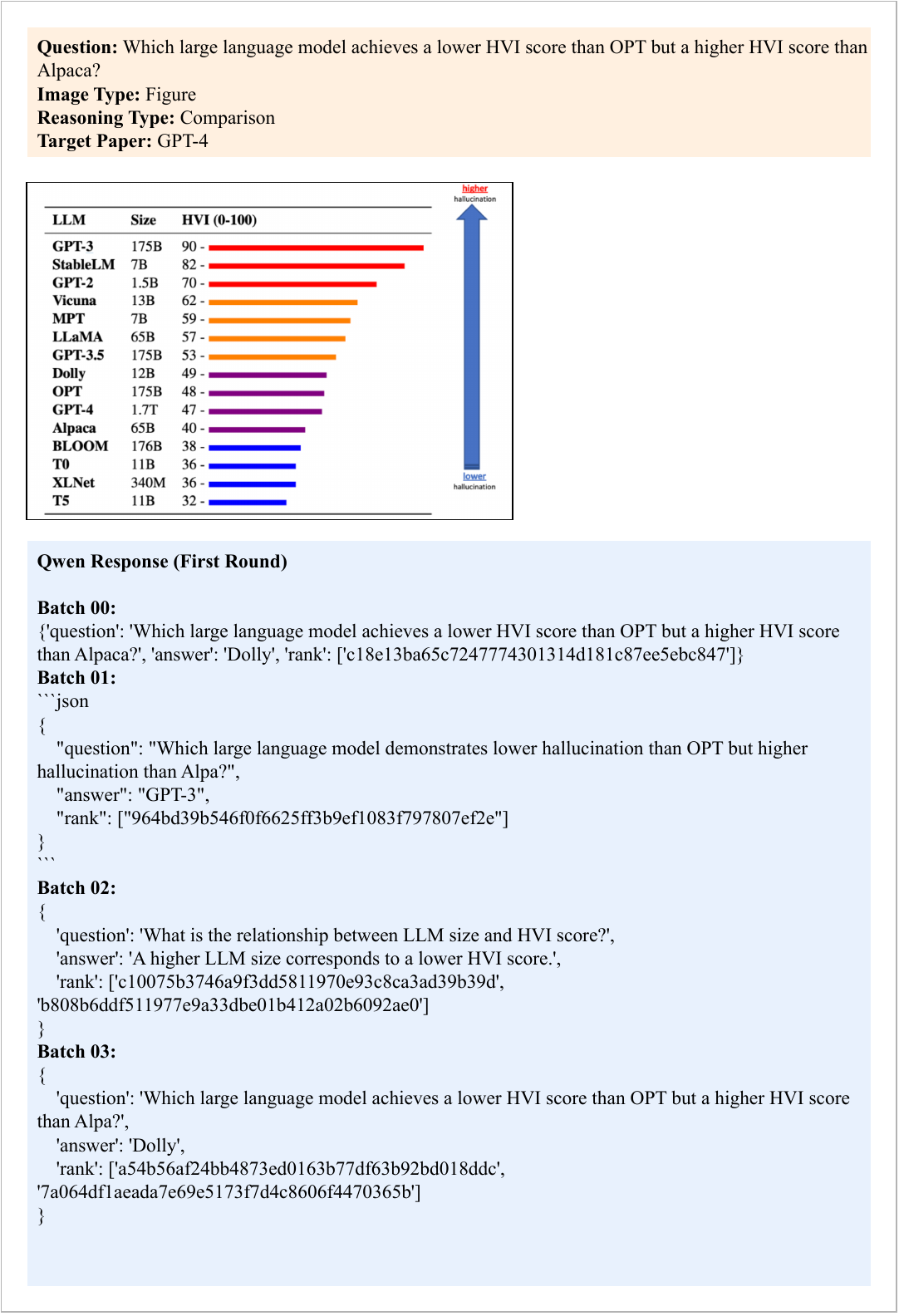}
    \caption{Qwen example output on visual context question - Part 1 of 3.}
    \label{fig:qwen_locality1}
\end{figure*}
\clearpage
\newpage
    
\begin{figure*}[p]
    \centering
    \ContinuedFloat
    \includegraphics[width=\textwidth]{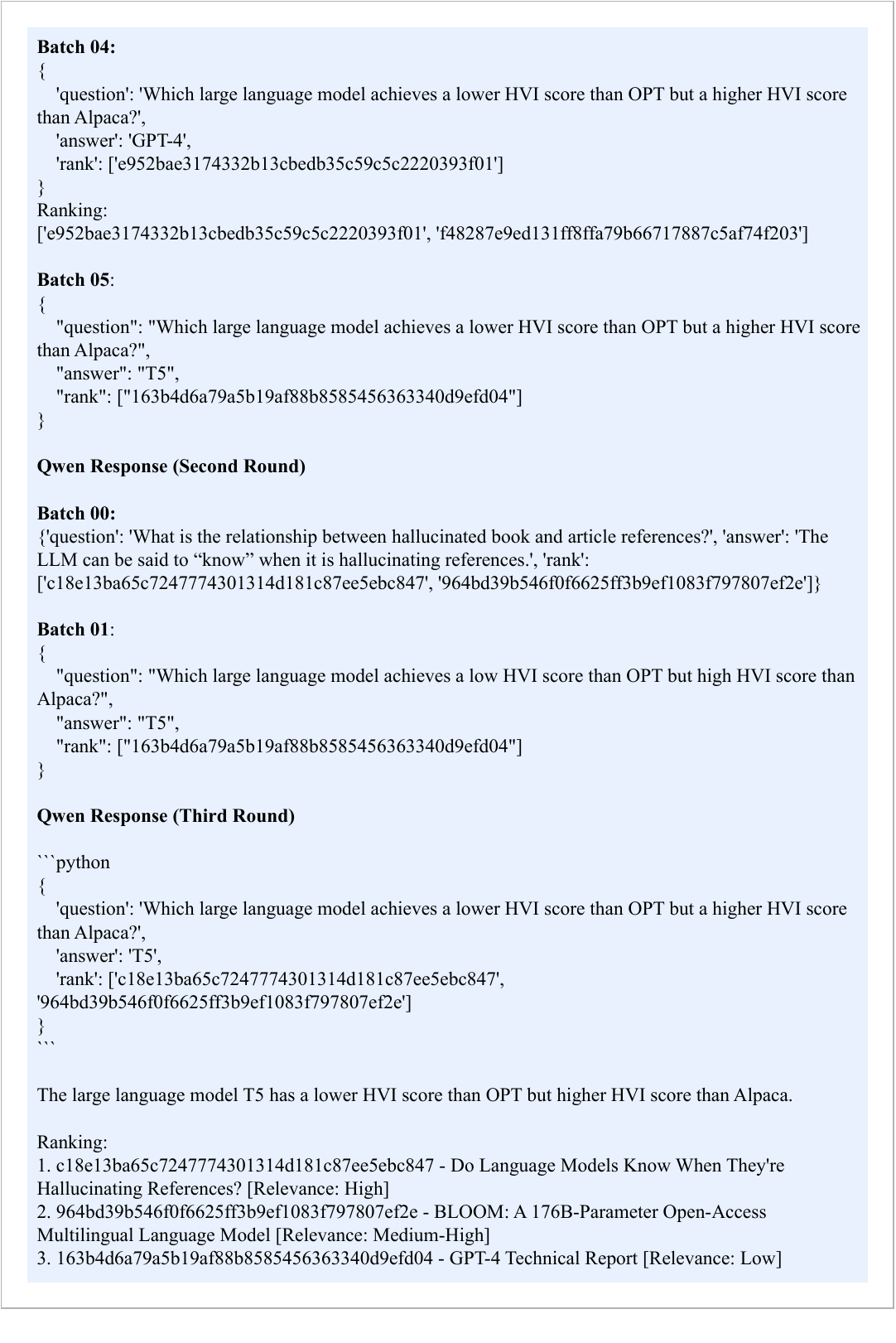}
    \caption{Qwen example output on visual context question - Part 2 of 3.}
    \label{fig:qwen_locality2}
\end{figure*}
\clearpage
\newpage
    
\begin{figure*}[p]
    \centering
    \ContinuedFloat
    \includegraphics[width=\textwidth]{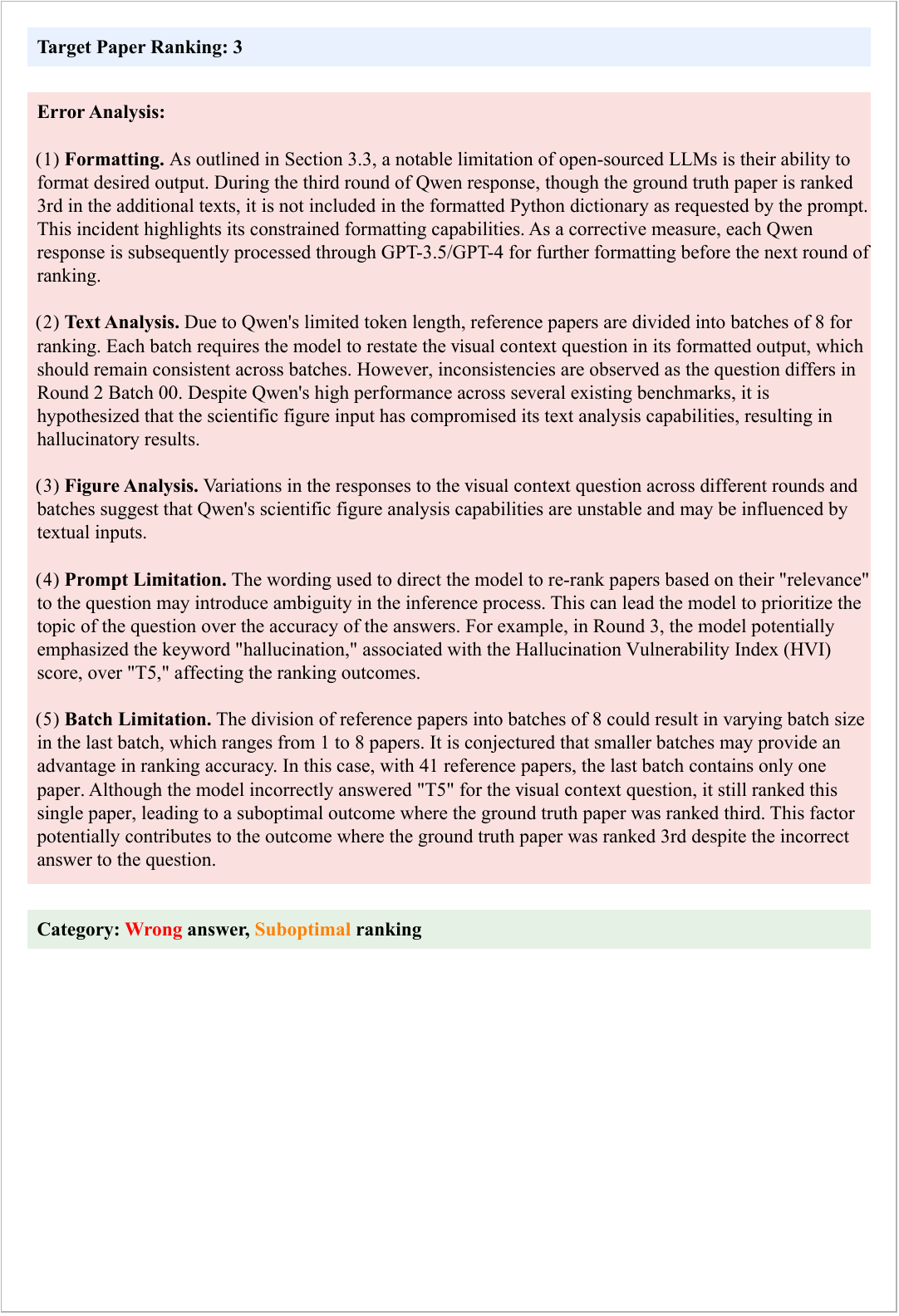}
    \caption{Qwen example output on visual context question - Part 3 of 3.}
    \label{fig:qwen_locality3}
\end{figure*}
\clearpage
\newpage

\begin{figure*}[p]
    \centering
    \includegraphics[width=\textwidth]{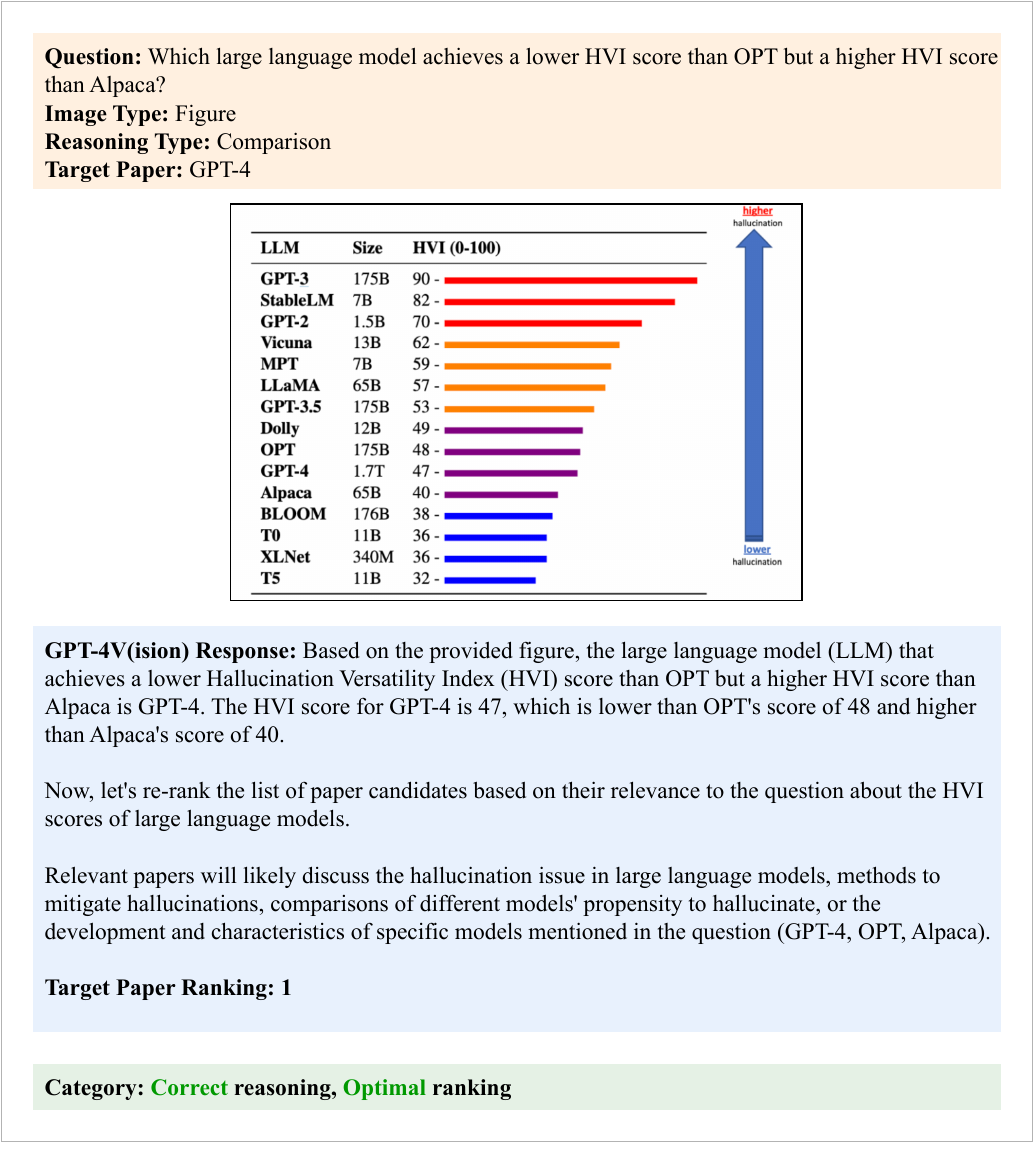}
    \caption{GPT-4V(ision) example output 1 on visual context question.}
    \label{fig:gpt_locality1}
\end{figure*}
\clearpage
\newpage
\begin{figure*}[p]
    \centering
    \includegraphics[width=\textwidth]{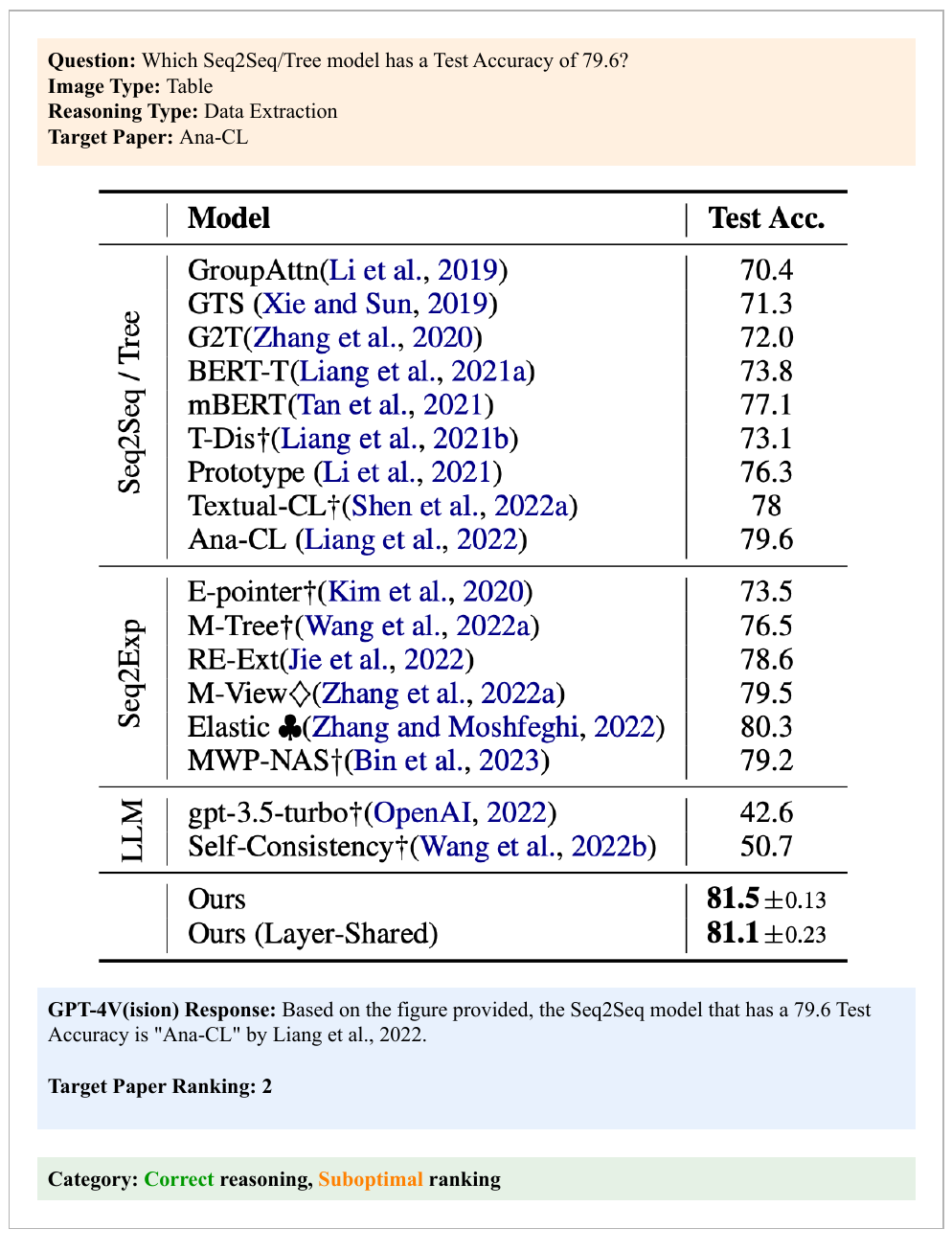}
    \caption{GPT-4V(ision) example output 2 on visual context question.}
    \label{fig:gpt_locality2}
\end{figure*}
\clearpage
\newpage

\begin{figure*}[p]
    \centering
    \includegraphics[width=\textwidth]{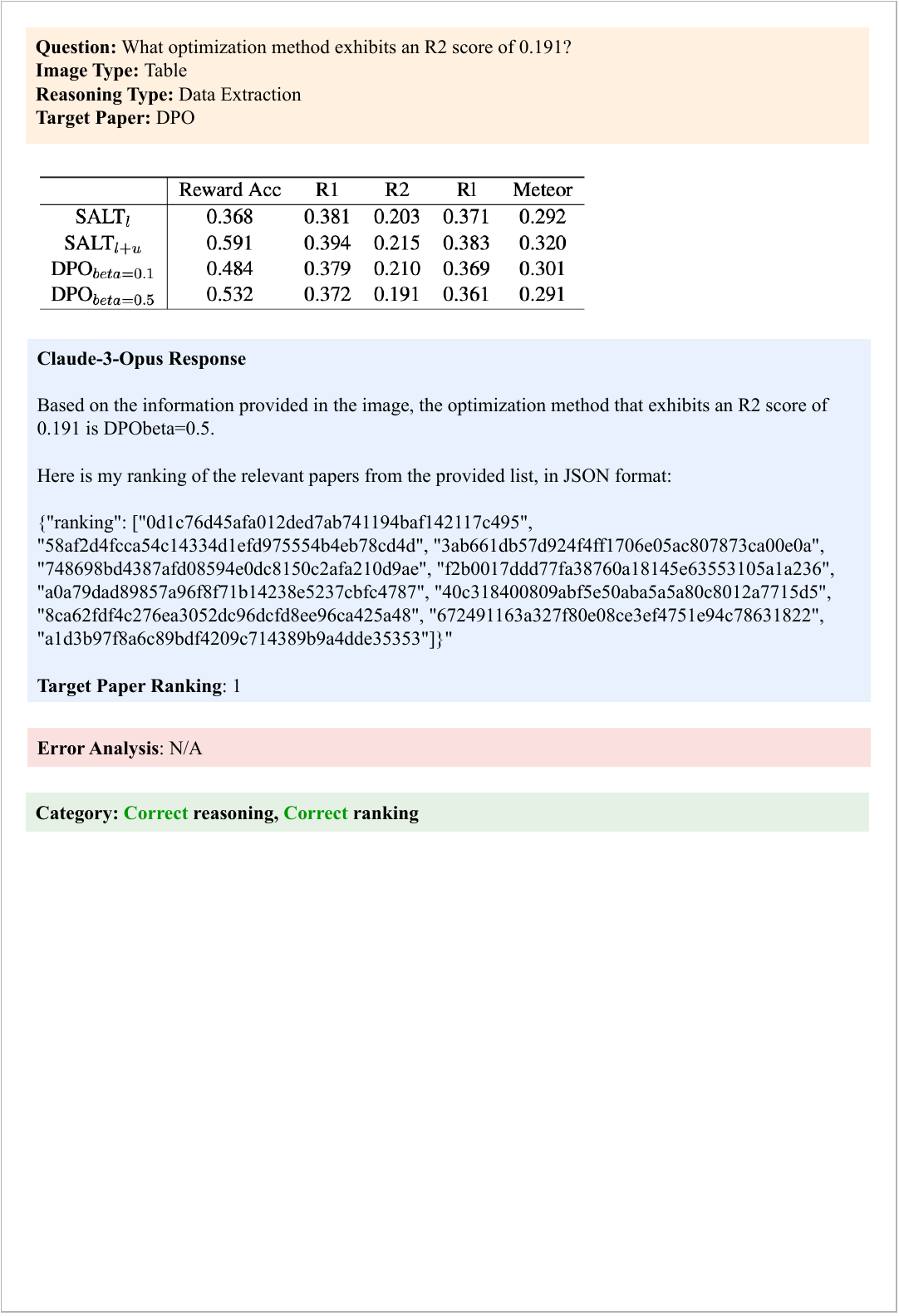}
    \caption{Claude-3-Opus example output 1 on visual context question.}
    \label{fig:opus_locality1}
\end{figure*}
\clearpage
\newpage

\begin{figure*}[p]
    \centering
    \includegraphics[width=\textwidth]{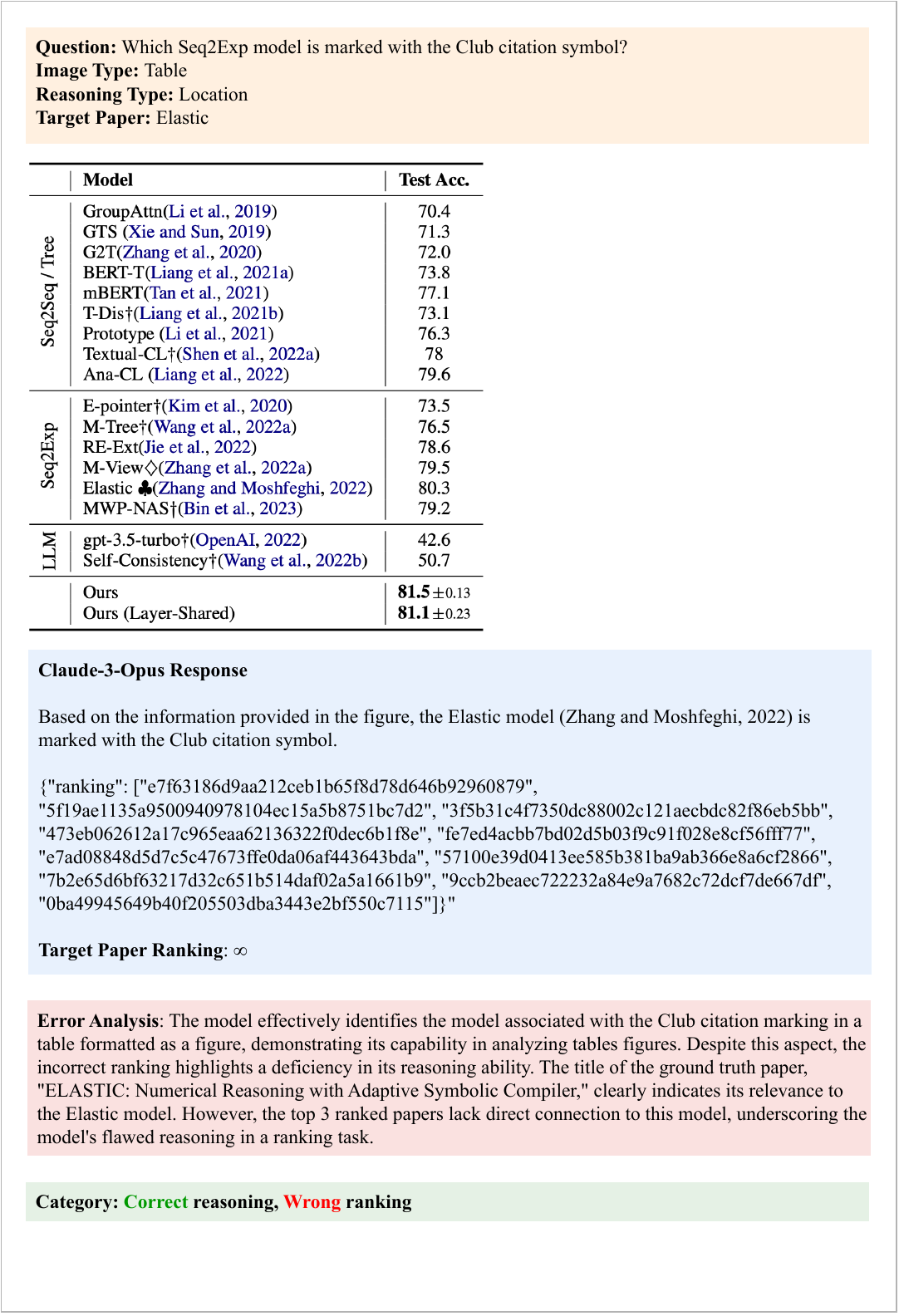}
    \caption{Claude-3-Opus example output 2 on visual context question.}
    \label{fig:opus_locality2}
\end{figure*}
\clearpage
\newpage

\end{document}